\documentclass[sigconf]{acmart}

\AtBeginDocument{%
  \providecommand\BibTeX{{%
    \normalfont B\kern-0.5em{\scshape i\kern-0.25em b}\kern-0.8em\TeX}}}

\setcopyright{acmcopyright}
\copyrightyear{2021}
\acmYear{2021}
\acmDOI{}

\acmConference[FAccT, March 3--10, 2021, Virtual Event, Canada]{}{}{}
\acmBooktitle{}
\acmPrice{}
\acmISBN{ACM ISBN 978-1-4503-8309-7/21/03}

\usepackage{amsmath}
\begin{document}

\title{Biases in Generative Art--- A Causal Look from the Lens of Art History}

\author{Ramya Srinivasan}
\affiliation{Fujitsu Laboratories of America}
\author{Kanji Uchino}
\affiliation{Fujitsu Laboratories of America}





\begin{abstract}
With rapid progress in artificial intelligence (AI), popularity of generative art has grown substantially. From creating paintings to generating novel art styles, AI based generative art has showcased a variety of applications. However, there has been little focus concerning the ethical impacts of AI based generative art. In this work, we investigate biases in the generative art AI pipeline right from those that can originate due to improper problem formulation to those related to algorithm design. Viewing from the lens of art history, we discuss the socio-cultural impacts of these biases. Leveraging causal models, we highlight how current methods fall short in modeling the process of art creation and thus contribute to various types of biases. We illustrate the same through case studies, in particular those related to style transfer. To the best of our knowledge, this is the first extensive analysis that investigates biases in the generative art AI pipeline from the perspective of art history. We hope our work sparks interdisciplinary discussions related to accountability of generative art.
\end{abstract}




\maketitle
\section{Introduction}
Generative art refers to art that in part or in whole has been created by the use of an autonomous system \cite{genart}. In a broad sense, the term ``autonomous" can refer to any non-human system that can determine features of an art work, such as use of smart materials, mechanical processes, and chemical processes, to name a few. Computer generated art, i.e. art generated by algorithms or computer programs, is perhaps the most common form of generative art \cite{wiki}. In fact, the terms "generative art" and "computer art" have been used more or less interchangeably for a long time now \cite{genart, wiki}.

With the rapid advancement of deep learning, there has been remarkable progress in AI based generative art. From creating hybrid images using attributes from multiple images to generating cartoons from portraits, AI based generative art has exemplified new and diverse applications  \cite{mazzone, gatys, artbreeder, deepdream}. 

As a consequence, AI based generative art has become extremely popular. In 2019, ``Sotheby" auction house sold an AI generated art work for 32000 pounds \cite{value}. A little earlier in 2018, the auction house ``Christies" sold an AI generated art work for a staggering 432500 USD \cite{christies}. Institutions have also shown an increased interest towards AI based generative art as evidenced by the number of museum shows \cite{history, elgammalnyc}. There are also several apps and tools such as \cite{artbreeder, aipainter, deepart} that have not only enhanced the popularity of AI based generative art, but have also facilitated ease of use and accessibility to end-users. 

\begin{figure}[h]
  \centering
   \includegraphics[width=0.25\textwidth]{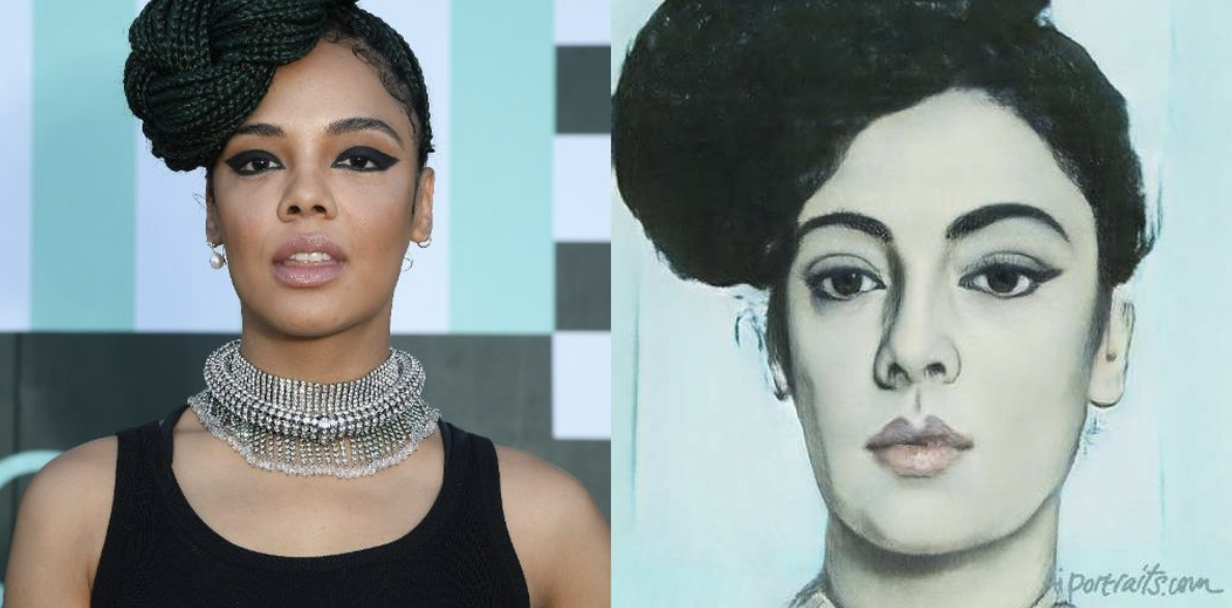}
    \caption{\small {Example of bias in AIportraits app \cite{aiportraits}---Skin color of actress Tessa Thompson (left) is lightened in the app's portrait rendition (right), thus exhibiting racial bias. Image source \cite{sung}.}}
\end{figure}

Amidst this progress and popularity surge, experts across disciplines have voiced concerns regarding the consequences of generative art. For example, in \cite{aies}, the authors reflect on the impact of interfacing with art autonomously generated by non-human creative agents. The authors argue that such art could widen the divide between the human creator and the human audience, and emphasize on the need to balance between automation and development of humans' creative potentials. Discussing whether computers can replace human artists, the author in \cite{hertzman} argues that {\it ``art requires human intent, inspiration, a desire to express something"}. The author further adds that artistic creation is primarily a social act, and concludes that computers cannot replace humans. In a similar vein, discussing whether machines can create art, the author in \cite{mark} argues that the process of creating an art work is distinct from the outcome, i.e. the artwork. Drawing from the theory of philosophical aesthetics, the author states that when something is created, something about inner self is expressed. 

It has been argued that both art and technology are means through which humans reveal their epistemic knowledge \cite{heidegger}. Such knowledge could include societal values, cultures, beliefs, as well as individual biases and prejudices. 
In the work \cite{hertzman}, artist and computer scientist Aaron Hertzman states {\it ``artworks are created by a human-defined procedure"}, and further notes that computer generated art can thus be biased. Even artists like David Young who argue that machines create art on their own, acknowledge existential bias in generative art. In the essay {\it Tabula Rasa} \cite{tabularasa}, Young notes that human biases in the form of preconceptions, irrationalities, and emotions can easily get embedded into the data used to train these generative art AI models. A recent notable example of bias in generative art concerns a portrait generator app called ``AIportraits" \cite{aiportraits}. It was pointed out that skin color of people of color was lightened in the app's portrait rendition \cite{vice, sung}. Figure 1 provides an illustration of the same. Furthermore,  AI algorithms are best thought of as data-fitting procedures \cite{mit}. As the author in \cite{hertzman} beautifully describes, these algorithms are {\it ``like tourists in a foreign country that can repeat and combine phrases from the phrasebook, but not truly understand the foreign language or culture"}.

\begin{figure}[t]
  \centering
    \includegraphics[width=0.5\textwidth]{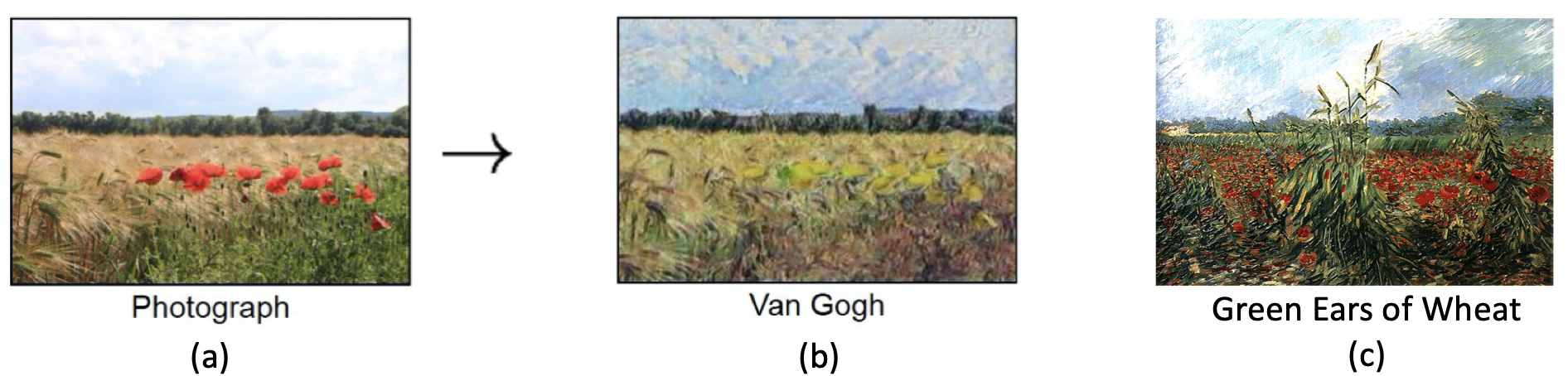}
    \caption{{\small Example of bias in learning artists' styles. The affect conveyed by the original image (a) is lost in the ``Van Gogh" version (b) of \cite{cyclegan}. This is contrary to Van Gogh's style, see (c), a real artwork by Van Gogh depicting red flowers in the field like in (a).}}
\end{figure}

Many types of latent biases in generative art, especially those concerning art history, have not been analyzed in any of the past studies. Further, socio-cultural impacts of biases in generative art have not been investigated. We aim to address these issues in this work. We motivate the problem with an illustration. Consider an image-to-image translation model such as CycleGAN \cite{cyclegan} which has been used to create images across different artists' ``styles". Figure 2 (b) shows an example of ``Van Gogh" version of a photograph (Figure 2(a)) as rendered by the CycleGAN model. As can be seen, the affect conveyed by the original image is lost in the ``Van Gogh" version: the red flowers, perhaps indicative of Spring are no longer evident, instead a dry season is reflected.

The generated image seems quite contrary to Van Gogh's take on colors. {\it ``Green Ears of Wheat"}, an 1888 art work by Van Gogh serves as an illustration to the point; see Figure 2 (c) \cite{wikiart}. As documented in his letters to his sister Wilhelmina and artist Horace M. Livens, Van Gogh mentions about how he emphasized colors \cite{gogh}, and how by using bright contrasting colors he was able to infuse life in his works: {\it `` Poppies or red geraniums in vigorously green leaves - motif in red and green. These are fundamentals, which one may subdivide further, and elaborate, but quite enough to show you without the help of a picture that there are colours which cause each other to shine brilliantly, which form a couple, which complete each other like man and woman”}, \cite{johanna,harrison}. Thus, by merely using correlation statistics to model the artist's style, aspects such as emotion and intent which are central to an art's creation \cite{mark} are not taken into account, thereby rendering a biased representation of the artist's ``style" and also possibly stereotyping the artist in the process.

Such biases could potentially have long standing adverse socio-cultural impacts. First, because of the inherent biases in training data and algorithms, generative art could be embedded with racial bias, gender bias, and other types of discrimination. Second, based on their limited understanding, algorithms could stereotype artists' style and not reflect their true cognitive abilities. This means aspects such as artist's intent and emotions are overlooked, thus potentially conveying an opposite affect in the generated art. Third, historical events and people may be depicted in a manner contrary to the original times, thus contributing to a bias in understanding history and thereby get in the way of authentically preserving cultural heritage (illustrated in Sec. 6). 
These observations therefore compel an analysis of generative art so as to uncover various types of biases. 
\subsection{Contributions}
In this paper, we investigate biases in AI based generative art right from those that can originate due to inappropriate problem formulation to those that can be related to algorithm design. Viewing from the lens of art history, we discuss the social-cultural impacts of these biases. We advocate for the use of causal models \cite{pearl} to depict potential processes of art creation. We highlight how current methods can fall short in modeling the process and thus contribute to various biases such as selection bias and transportability bias \cite{elias}. We illustrate the same through case studies that span various art movements, artists, art media/material, genres, and geographies. 

First, examples of biases that arise due to improper problem formulation, namely, framing effect biases, are considered (Sec. 4).  After providing a background of causal models (Sec. 5), we discuss case studies to demonstrate confounding biases in modeling artists' styles (Sec. 6.1), followed by illustrations of selection bias (Sec. 6.2). Biases in transferring artists styles, also known as transportability biases, are discussed (Sec. 6.3). Transportability bias is demonstrated by considering case studies that include both art movements that are subtly different (e.g. modern art movements such as Cubism and Futurism) and art movements across different time periods (e.g. Modern art and Early Renaissance). Illustrations of biases in datasets are provided in Sec. 7.  A list of the case studies examined in the paper can be found in Table 1. Our findings shed light on the various inherent biases in generative art right from problem formulation and datasets to algorithm design and data analysis. We also discuss the socio-cultural repercussions of these biases (Sec. 8). To the best of our knowledge, this is the first extensive analysis that investigates biases in the generative art AI pipeline from the perspective of art history. We hope our work triggers interdisciplinary discussions concerning accountability of generative art, and sparks the design of novel methods to address these issues. 
\begin{table*}
\begin{tabular}{|l|l|l|l|l|l|}
\hline
{\bf CS}& {\bf Model} &  {\bf Bias type} &  {\bf Art Movements} &  {\bf Artists} &  {\bf Genres} \\
\hline
1&\cite{cyclegan}  &  Confounding bias & Post-Impressionism & Vincent Van Gogh & Landscape\\ \hline
2& \cite{artgan}  &  Selection bias & Romanticism & Gustave Dore & Illustration\\ \hline
 3& \cite{aiportraits}  &  Selection bias (racial bias) & Renaissance & Various artists & Portraits\\ \hline
 4& \cite{cyclegan}  &  Transportability bias  & Post-Impressionism & Paul Cezanne & Photo, Landscape \\ \hline
5& \cite{deepart}  &  Transportability bias & Cubism, Futurism & Fernand Leger, Gino Severini & Genre art, battle painting \\ \hline
6& \cite{deepart}  &  Transportability bias & Realism, Expressionism & Mary Cassatt, Ernst Kirchner & Portraits\\ \hline
7& \cite{goart}  &  Transportability bias & Renaissance, Expressionism, & Clementine Hunter, & Portrait, Sculpture\\ 
&  & (racial bias)  & Folkart & Desiderio da Settignano & \\ \hline
8& \cite{abacus} & Transportability (gender) bias  & Renaissance & Raphael, Piero di Cosimo & Portraits \\ \hline
9& \cite{aiportraits} & Representational bias & Renaissance & Various artists & Portraits \\ \hline
10& \cite{artgan} & Label bias & Ukiyo-e & Various artists & Various genres \\ \hline
\end{tabular}
\caption{{\small Summary of Case Studies (CS) described in the paper. Note, there could be more than one type of bias associated with each CS. For illustrative purposes, only one bias type is discussed in each CS. Model denotes the  algorithm/platform listed under corresponding references.}}\label{tab1}
\end{table*}
\subsection{Case study selection}
We surveyed academic papers, online platforms, and apps that generate art using AI. In order to uncover potential biases from an art historical perspective, from the surveyed list, we selected papers and platforms that focused on simulating established art movements and/or artists' styles. Thus, papers such as \cite{mazzone} or platforms such as \cite{artbreeder} which focus on deviating from established styles to create imaginary patterns are not included in our study. To demonstrate various biases, we have considered state-of-the-art generative art AI models \cite{cyclegan, artgan} and platforms/apps such as \cite{deepart,aiportraits,goart} that focus on simulating established art movements and artists' styles. The art movements considered as part of case studies have been determined based on the experimental set-ups reported in these state-of-the-art AI models and platforms. These include Renaissance art, Modern art (Cubism, Futurism, Impressionism, Expressionism, Post Impressionism and Romanticism), and Ukiyo-e art. It is to be noted that due to the paucity of existing AI applications that study non-western art forms, Ukiyo-e is the only non-western art form studied. The genres span landscapes, portraits, battle paintings, genre art, sketches, and illustrations. Art material considered includes woodblock prints, engravings, paint, etc.  The study includes artists across cultures such as Black folk artist Clementine Hunter, American painter Mary Cassatt, Dutch artist Van Gogh, French illustrator and sculptor Gustave Dore, Italian artist Gino Severini, amongst others. In the next section, we  discuss work related to computer generated art.
\section{AI for Art Generation}
Computer generated art has a long history. In 1970’s, painter Harold Cohen began exhibiting paintings generated by a program called AARON \cite{hertzman}. By 1980s, several artists were using computer programs to create interactive experiences for the audience \cite{hertzman}. In the 1990s, Flash, a tool for creating animations became popular \cite{history}. Around the same time, Paul Haeberli introduced a paint program whereby a user could quickly create a painting without needing any technical skill \cite{paul}. In 2000s, tools like Processing \cite{reas} and OpenFrameworks \cite{liberman} allowed artists to make art using code. 

As early as 2001, researchers in computer vision were training computers to learn artists' styles from examples \cite{jacobs}. Since 2012, rapid advancement in deep learning has triggered a wide range of models for AI based generative art. For example, \cite{deepdream1,  miller} is an open source tool released by Google that uses a convolutional neural network (CNN) to understand what neural networks are doing at each layer by creating dream-like appearances. Another CNN architecture is the neural style transfer \cite{gatys} work that allows to blend content of one image into style of another image to create new images. In \cite{magenta}, a recurrent neural network is proposed to construct stroke-based drawings of common objects. 

Recently, generative adversarial networks (GANs) \cite{goodfellow} have become popular for creating art. Creative adversarial networks \cite{mazzone} proposed modifications to the GAN objective to make it creative by maximizing deviation from established styles and minimizing deviation from art distribution. In \cite{cyclegan}, the authors proposed a method to translate styles across unpaired images and illustrated its applicability in transferring artists' styles. In \cite{artgan}, the authors used conditional GANs to generate artworks.

There are many tools and apps to facilitate users to easily create art. Using \cite{aiportraits}, users can transform a portrait in the style of famous portraits. Photos can be converted into artworks using \cite{aipainter}. In an application called GANbreeder, two images are combined to generate a new image \cite{artbreeder}. The style of the input image is transformed into another specified style in \cite{deepart}.  Cartoonify turns a photo into a cartoon drawing leveraging Google's ``Draw This" \cite{drawthis}. 

Given that there are many tools to quickly create art, there is an increased risk for generative art to be biased. With surging art market \cite{market} and pressing need for diversity and inclusiveness in art \cite{predict}, an analysis of bias in generative art becomes even more pertinent. In a recent art project titled {\it `Imagenet Roulette'} \cite{kate}, AI researcher Kate Crawford and artist Trevor Paglen exposed biases in machine learning datasets. While there have been studies to understand how humans perceive art \cite{hong} and to examine if there is a perception bias towards art created by AI \cite{ragot}, there is little to no extensive analysis concerning biases {\it in} generative art. In this paper, we provide an extensive analysis of biases in art generated autonomously by AI, looking from the perspective of art history. Such an analysis is useful to understand machine related biases independent of human biases, and is necessary to study scenarios involving both human (artist) and machine bias. In the rest of this paper, generative art refers to AI based generative art. 

\section{Art-historical aspects of artworks}
In this section, we discuss various aspects pertaining to art history based on which art works can be analyzed. Art historians employ a number of ways to group world arts into systems of classification \cite{wikiart}. These groupings are based on a set of qualities that are significant. Such significant qualities could be related to specific approach of an artist, material used to create art works, art movements, genre, etc. We provide a brief account of some main aspects so as to aid in understanding some types of biases that we illustrate in the paper. 
\subsection{Art Movements} 
Art movements can be described as tendencies or styles in art with a specific common philosophy influenced by various factors such as cultures, geographies, political-dynastical markers, etc. and followed by a group of artists during a specific period of time \cite{wikiart}. Examples of art movements include Ancient Egyptian art, Ancient Greek art, Medieval Art, Renaissance art, Modern art, etc. Within each of these art movements, there are sub-categories based on various factors. For example, modern art includes many sub-categories such as Symbolism, Impressionism, Post-impressionism, Cubism, Futurism, Pop-art, and so on. As an illustration, consider Impressionism and  Post-impressionism. Both movements originated in France, however Post-impressionism originated in reaction to Impressionism. While Impressionism was characterized by vibrant colors, spontaneous brush strokes, and urban life styles, Post-Impressionism artists had their own individual styles to symbolically display real subjects and their emotions \cite{oxford}. Similarly, each art movement is characterized by unique features that reflects certain trends. Thus art movement is a dominant factor influencing artists and artworks. Interested readers may refer to \cite{wikiart} where over hundred sub-categories are listed across a dozen art movements. 
\subsection{Art material}
Artworks can also be grouped based on the material and techniques used in creating the art.  Charcoal, enamel, mosaics, tapestry, paint, and lithography are some examples of art materials. Artists use different techniques to create artworks from different materials. For example, mosaic is a coherent pattern or image in which each component element is built up from small regular or irregular pieces of substances such as stone, glass or ceramic, held in place by plaster/mortar, entirely or predominantly covering a plane or curved surface, even a three dimensional shape, and normally integrated with its architectural context \cite{mosaic}. Mosaics were traditionally used as decoration for floors and walls becoming very popular across the Ancient Roman World. Different art movements saw the prevalence of different materials. For example during the Renaissance period, sculptures were made out of various materials like marble, white stone, gold, etc. Thus, material and technique employed to create art can influence the artist and the resulting artwork. An elaborate list of various materials can be found in the WikiArt dataset \cite{wikiart}.  
\subsection{Genre}
Genre of an artwork is based on the depicted themes and objects. A hierarchy of genres was developed in the $17^{th}$ century  \cite{canon}. According to this hierarchy, history paintings, namely paintings depicting scenes of important historical, mythological, and religious events, were considered to be the top ranked genre and this was so until the mid $19^{th}$ century. History paintings were usually large and typically narrated a story such as a battle, allegory, or the like.  Portraiture was another prominent genre and these usually depicted royals, aristocrats, and other important people in society. Portraiture had to convey aesthetic aspects of the subject depicted, such as their power, beauty, etc. In contrast, ``genre painting" depicted scenes from every day lives of ordinary people. Landscapes, animal painting, and still life painting are some other prominent genres. Abstract or figurative art are the most common genres for contemporary art \cite{wikiart}. An artist usually can work across different genre types, however, art historians mark certain artists as representatives of a particular genre. For example, Anthony van Dyck is recognized as a portraitist, Alfred Sisley as a landscape painter, and Piet Mondrian as an abstract artist – though each one of them worked in a number of different genres \cite{wikiart}. Wikiart dataset provides an extensive list of genres based on artists and artworks.
\subsection{Artists}
There are many aspects that characterize an artist's  ``style". In addition to factors such as art movement, art material, and genre, an artist's style can be characterized by factors such as their cultural backgrounds, their art lineage or schools (from whom they learned or who influenced them), and other subjective aspects such as their cognitive skills, beliefs, prejudices, and so on. Consider for example, Paul Cezanne, one of the most popular artists in the history of modern art. Although generally categorized as a Post-Impressionist artist, Cezanne influenced several other art movements such as Cubism and Fauvism. Some of his early pictures depict classical and romantic themes with expressive brushwork and dark colors, while later he is said to have adopted brighter colors drawing inspiration, emotions, and memory to paint \cite{james}. Cezanne himself remarked: {\it ``A work of art which did not begin in emotion is not art"}. In his still-life paintings, Cezanne began to address technical problems of form and color by experimenting with subtly gradated tonal variations, or “constructive brushstrokes,” to create dimension in the objects \cite{james}.  His artworks span a variety of genres and exhibit patterns from multiple art movements marked by subtle cognitive aspects. Thus, modeling  artists' style is not a straightforward computational task, it entails many abstract elements that are hard to quantify. Yet, researchers define ``style" in ways that suit their model's performance, we discuss this issue next.
\section{Biases due to Problem Formulation}
Biases can arise based on how a problem is defined and is formally known as the {\it framing effect bias} \cite{plous}. Consider, for example, the problem of style transfer in artworks. There are at least two different notions of styles when it comes to artworks: one which is related to the art movement, and another which is related to the artist. As described in Section 3, there are many aspects to each of the above notions of style. Yet researchers conveniently define style in a manner that suits their model's performance. This is a consistent problem across several models. For example, in \cite{artgan}, the authors claim that their model learns {\it Ukiyo-e} style since the generated images are ``yellowish" like Ukiyo-e artworks. In \cite{cyclegan}, a single model is used to learn styles of artists and art movements. Like in \cite{artgan}, the justification to have learned Ukiyo-e style seems to be based on color features. Thus, ``style" has been defined based on the color of the generated art. Given that most Ukiyo-e works were woodblock prints, it is thus natural for the generated art to be yellowish.  

Ukiyo-e is a form of Japanese art. The works usually depicted landscapes, tales from history, scenes from the Kabuki theatre, and other aspects of everyday city life. Some unique characteristics of Ukiyo-e included exaggerated foreshortening, asymmetry of design, areas of flat (unshaded) colour, and imaginative cropping of figures \cite{ukiyo}. Foreshortening refers to the technique of depicting an object or human body in a picture so as to produce an illusion of projection or extension in space and to convey the notion of depth. These characteristics are not captured in the examples depicted in \cite{artgan, cyclegan}. Such drawbacks can also happen due to the model design issues which we discuss in Section 6. Nevertheless, inappropriate problem formulation can introduce and perpetuate biases across the generative art AI pipeline. 
In the next section, we briefly describe structural causal models that we leverage to analyze different biases. 
\section{Structural Causal Models}
We advocate for the use of causal directed acyclic graphs (DAGs) \cite{pearl} to analyze some types of biases that are related to algorithm design and datasets. In Section 3, we saw that there are several aspects relevant to an artwork and that these aspects could influence each other in many ways. For example, art movement could influence the choice of art material, the subject of a portrait could influence the artist, and so on. DAGs help in visualizing such relationships. DAGs also allow encoding of assumptions about data, model, and analysis, and serve as a tool to test for various biases under such assumptions. Researchers have leveraged causal models to discuss and develop various notions of fairness \cite{glymour, silva}. Causal models facilitate domain experts such as art historians to encode their assumptions, and hence serve as {\it accessible data visualization and analysis} tools. Based on different expert opinions, there can be multiple assumptions. Thus, there can be more than one DAG describing the relationship of an artwork with the artist, genre, art movement, etc. Thus, using multiple DAGs it is possible to analyze for biases under different scenarios. We discuss basic concepts related to causal models, and through case studies, illustrate the intuition behind using causal models to analyze biases in generative art.

DAGs are visual graphs for encoding causal assumptions between the variables of interest. Specifically, the assumptions about data are encoded by means of structural causal models (SCMs) \cite{pearl}. A structural causal model $M$, consists of two sets of variables, $U$ and $V$, and a set $F$
of functions that determine or simulate how values are assigned to
each variable $V_{i} \in V$. 
The variables $V$ are observed and variables $U$ are unobserved.  Variables $U$ and $V$ constitute the vertices of a causal graph $G$ and the directed edges between them denote the various causal dependencies.
\subsection{d-separation} Regardless of the functional form of the equations
in the model ($F$), conditional independence relations can be obtained if the model $M$ satisfies certain criteria. $d-separation$ is a criterion for deciding, from a given a causal graph, whether a set $X$ of variables is independent of another set $Z$, given a third set $Y$. The idea is to associate ``dependence" with ``connectedness" (i.e., the existence of a connecting path) and ``independence" with ``unconnected-ness" or ``separation" \cite{pearl}. Path here refers to any consecutive sequence of edges, disregarding their direction.
\begin{figure}[t]
  \centering
    \includegraphics[width=0.45\textwidth]{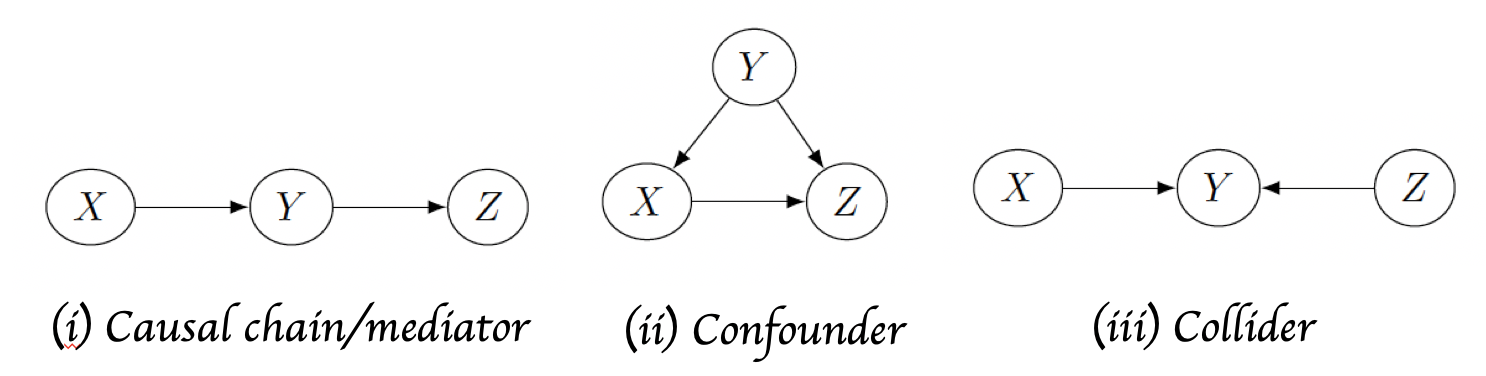}
    \caption{{\small D-separation: Graph structures to illustrate conditional independence. Please refer to Section 5.1 for details} }
\end{figure}

Consider a three vertex graph consisting of vertices $X$, $Y$, and $Z$. There are three basic types of relations using which any pattern of arrows in a DAG can be analyzed, these are depicted in Figure 3. The leftmost graph (i) denotes a causal chain or that of a ``mediation". The effect of $X$ on $Z$ is mediated through $Y$. In this case, given $Y$ (i.e. conditioning on $Y$), $X$ is independent of $Z$ or $Y$ is said to $block$ the path from $X$ to $Z$. In the center graph (ii), $Y$ is a common cause of $X$ and $Z$. If unobserved, $Y$ is a confounder as it causes spurious correlations between $X$ and $Z$. Conditioned on $Y$, the path from $X$ to $Y$ is blocked. In the rightmost graph (iii), $Y$ is a collider as two arrows enter into it. As such, the path from $X$ to $Y$ is blocked. However, conditioning on $Y$, the path will be unblocked. In general, a set $Y$ is admissible (or “sufficient”) for estimating the causal effect of $X$ on $Z$
if the following two conditions hold \cite{pearl}:
\begin{itemize}
{\item No element of $Y$ is a descendant of $X$ }
{\item The elements of $Y$ block all backdoor paths from $X$ to
$Z$—i.e., all paths that end with an arrow pointing to $X$.}
\end{itemize}
\subsection{Interventions}
An $intervention$ on a graph is denoted by the ``$do$" operator \cite{pearl}. The $do$ operator corresponds to setting the intervened variable to a specific value and removing the influence of other variables on it. For example, in graph (ii) of Figure 3, $do(X=x)$ implies that the variable $X$ is set to the value $x$ and the incoming arrow into $X$ is removed. $do$ operator facilitates quantification of causal effects. For example, in Figure 3 (ii), the expression $P(Z|do(X=x))$ quantifies the causal effect of $X$ on $Z$. A set of rules referred to as $do-calculus$ are always applicable in the context of interventions \cite{pearl}. These rules determine when it is possible to
\begin{itemize}
{\item to add/delete observations in interventions,}
{\item to exchange interventions and observations,}
{\item to add/delete interventions}
\end{itemize}
$do-calculus$ helps in rendering expressions free from $``do"$. We are now equipped to discuss algorithmic biases in generative art.
\section{Biases related to Algorithm}
Bias can arise if the algorithm ignores the effect of unobserved variables, overlooks domain specific differences, uses subsets of the population for analysis, and so on. We discuss these in this section.
\subsection{Confounding bias}
Confounding bias originates from common causes that affect both inputs and outputs \cite{pearl}. We illustrate this bias through case studies.
\subsubsection{Case Study 1:} Modeling artists styles has been one of the most common applications in generative art. As discussed in Section 3.4, several observed and unobserved abstract aspects constitute an artist's style. Revisiting example discussed in Figure 2, the problem of modeling Van Gogh's style can be viewed as estimating the causal effect of Van Gogh on the artwork. Thus, the expression $P(Art work|do(Artist))$ models the style of the artist in the artwork (Section 5.2).
It is to be noted that the assumptions encoded through a DAG can vary from one expert opinion to another. However, these varying opinions help to discover and test for biases under different scenarios and in turn highlight the drawbacks of existing correlation based methods such as \cite{cyclegan} that do not take into account important cultural and social aspects that influence an artist's style. 

Consider Figure 4 (i) that depicts one potential process of art creation (note there could be several others based on assumptions of domain experts, we consider one such for illustration). Here, the variable $X$ denotes the artist, $Z$ denotes the artwork, $G$ is the genre, $M$ is the art material, and $A$ denotes the art movement. According to the assumptions encoded in this DAG, art material, genre, and art movement are confounders influencing both the artist and the artwork. Further, art movement influences the art material. Let us assume that all of the confounders are observable. Under these assumptions, in order to model the style of Van Gogh (i.e. $P(Z|do(X=x))$), we have to block the backdoor path from $X$ to $Z$ (Section 5.1), so as to remove confounding bias. 

Specifically, for graph (i) in Figure 4, the following equation captures the causal effect of $X$ on $Z$ 
\begin{multline}
    P(Z|do(X=x))= \\ \sum_{g,a,m} P(Z|x,G=g,A=a,M=m )
    P(G=g,A=a,M=m)
    \end{multline}
For the case study, $do(X=x)$ implies {\it do(Artist=Van Gogh)}. The summation $\sum_{g,a,m}$ in Eq. 1 indicates that one has to consider all possible art movements, art materials, and genres that the artist has worked in order to model their style. The implication of finding a sufficient set, $A,G,M$, is that stratifying on $A,G,M$ is guaranteed to remove all confounding bias relative to the causal effect of $X$ on $Z$. 

Thus, modeling artist's style requires knowledge about the causal process governing the artwork's creation. Models like \cite{cyclegan, artgan} that do not consider the influence of confounders like art movement are prone to {\it omitted variable bias} \cite{wooldridge} and confounding bias. Art movements were characterized by many socio-cultural and political events and reflect the dynamic influence of these events on art. Thus by ignoring this variable, there is a bias in capturing the artist's style, in understanding the art's intent, and in representing culture. Based on the DAG, the causal effect may or may not be identifiable.  Eq. (2) differs from the conditional distribution $P(Z=z|X=x)$, and the difference between the two distributions, i.e. $P(Z|do(X=x))- P(Z=z|X=x)$ defines confounding bias \cite{elias}. 

Figure 4 (i) depicted a scenario with no unobserved confounders. In presence of unobserved confounders, causal effects are not identifiable. Consider Figure 4 (ii). Let $E$ represent emotions of the artist. The dotted circle and lines denote the unobserved confounder $E$ and its influence on other variables. In this case, even knowing the joint distribution of genre, art movement, and material $P(A,G,M)$ does not help in identifying Van Gogh's style, i.e. $ P(Z|do(X=x))$ is not identifiable. In general several subjective factors like prior beliefs, emotions, cultural values, etc. can influence the artist and the artwork. Thus in reality, the true style of any artist cannot be accurately modeled. Even if confounding bias due to unobserved factors is overlooked, there are other biases as discussed next.
\begin{figure}[t]
  \centering
    \includegraphics[width=.45\textwidth]{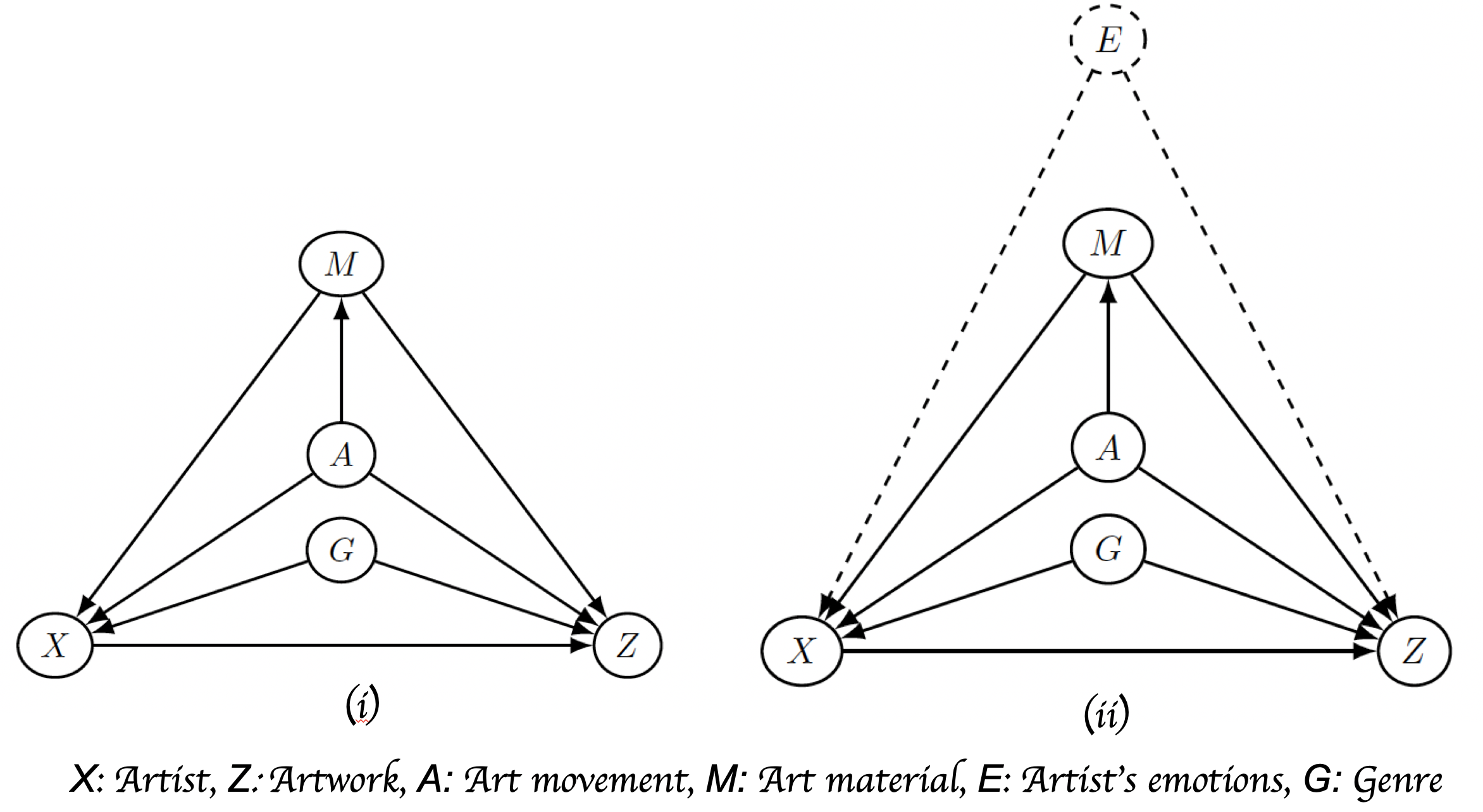}
    \caption{{\small (i): Confounding bias due to genre, material and art movements. (ii): Confounding bias due to artist's unobserved emotions.}}
\end{figure}

\subsection{Sample Selection bias}
Sample selection bias (or selection bias for short) is the bias that is induced by preferential selection of units for data analysis \cite{elias}. In a DAG, a special variable $S$ is used to denote the selection of the variable in the analysis, $S=1$ indicating selection and $S=0$ indicating otherwise. Consider for example, Figure 5 (i). Here, $X \rightarrow S$ and $Z \rightarrow S$ indicates that both inputs and outputs are selection dependent (i.e. $S=1$ with respect to both $X$ and $Z$). The case study discussed below will further clarify these concepts.

\subsubsection {Case Study 2} To illustrate selection bias, let us consider an example described in the ArtGAN model \cite{artgan}. The authors state that their model is able to recognize artist Gustave Dore's preferences as the generated images resonate with the dull color found in Dore's artworks. Mostly engravings were selected for analysis. Graph (ii) in Figure 5 depicts a possible DAG for this case.  Let $X$ denote Dore's style. $X \rightarrow S$ depicts the selection of engravings in the analysis. Further, as the authors mention, the generated images are greyish due to the engravings considered, thus $Z \rightarrow S$, where $Z$ denotes generated image. Additionally, there may be some unobserved confounders that influence both $X$ and $Z$ as denoted by the bi-directional dotted arrow, and there may be some other Artgan model variable $V$ that influences the nature of generated image $Z$, these are depicted in Figure 5 (ii). 
\begin{figure}[t]
  \centering
    \includegraphics[width=0.4\textwidth]{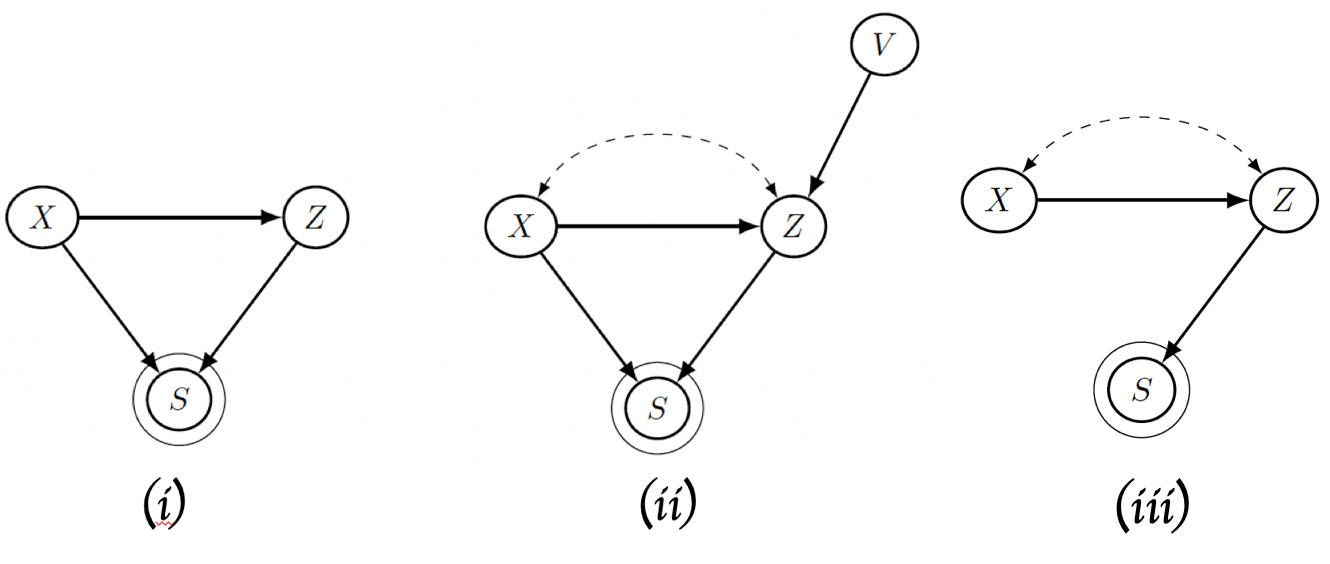}
    \caption{{\small (i): Example of selection bias (ii): Illustration of CS 2, see Section 6.2. (iii) Illustration of selection bias in datasets, see Section 7.1.  Causal effect of $X$ on $Z$ is not identifiable across (i), (ii), and (iii)}}
\end{figure}

In order to be able to recover the causal effect of $X$ on $Z$ under selection bias, \cite{elias} lists the conditions known as {\it selection backdoor criteria}. Formally, let $Y$ be a set of variables partitioned into two groups $Y^+$ and $Y^-$ such that $Y^+$ contains all non-descendants of $X$ and $Y^-$ the descendants of $X$, and let $G_s$ stand for the graph that includes the selection mechanism $S$. $Y$ is said to satisfy the {\it selection backdoor criterion} if the following conditions are true:
\begin{itemize}
{\item (i) $Y^+$ blocks all backdoor paths from $X$ to $Z$ in $G_s$ } 
{\item (ii) $X$ and $Y^+$ block all paths between $Y^-$ and $Z$ in $G_s$}
{\item (iii) $X$ and $Y$ block all paths between $S$ and $Z$ in $G_s$ }
{\item (iv) $Y$ and $Y \cup (X, Z)$ are measured.}
\end{itemize}
In Figure 5 (ii), the path between $S$ and $Z$ is not blocked due to the presence of direct link between the two variables, thus condition (iii) in selection backdoor criteria is not satisfied. Hence Dore's style cannot be recovered based on the engravings considered in the analysis. 
In fact, in addition to engravings, Dore's worked on paintings.  For example, landscapes such as {\it `The Lost Cow'}, paintings such as {\it `Little Red Riding Hood'}, religious painting such as {\it `The Wrestle of Jacob'} are not greyish and exhibit colors reflective of the genre.  Thus, by merely selecting engravings for analysis, sample selection bias is induced and style of Dore cannot be identified. Note, the failure to capture Dore's style in \cite{artgan} can be additionally attributed to other types of biases such as confounding bias, label bias, and even framing effect bias. For the purpose of illustrating selection bias, we have focused on describing selection bias only. In general, there can be more than one type of bias in a case study. 

\subsubsection{Case Study 3:} As an other example of selection bias, consider the case of \cite{aiportraits}. As illustrated in Figure 1, racial bias was evident in this application. Figure 5 (ii) can also be used to describe this scenario. Let $X$ denote the set of input images selected for analysis and let $Z$ represent the generated images. Selection of Renaissance portraits of mostly white people was used in the analysis, thus $X \rightarrow S$. As evident, the generated images were portraits of fair skinned people, thus $Z \rightarrow S$. Additionally, there could be unobserved confounders influencing both the input images $X$ and generated images $Z$, and there may be model parameters $V$ influencing the generated images. As condition (iii) in selection backdoor criteria is not satisfied, there is selection bias.
\subsection{Transportability bias}
Style transfer is a popular application in generative art. Various works have demonstrated transferring across artists' styles (e.g. Cezanne to Monet) and across art media (e.g. photograph to painting). Learning models that can generalize across domains is commonly known as transfer learning in the deep learning community and as transportability in the causality community. Transportability defines the conditions under which causal effects learned in experimental studies can be transferred into to a new population in which only observational studies can be conducted \cite{transport}. The differences between the populations of interest are expressed through representations called as ``selection diagrams”.  To this end, DAGs are augmented with a set, $R$, of “selection variables,” where each member of $R$ corresponds to a mechanism by which the two populations differ, and switching between the two populations will be represented by conditioning on different values of these $R$ variables. $R$ variables locate the mechanisms where
structural discrepancies between the two domains are suspected to take place. Transportability bias arises if causal effects cannot be transferred across populations.

The conditions under which causal effects can be transported are listed in \cite{aaai}. Formally, let $D$ be the selection diagram characterizing the two populations $\Pi$ and $\Pi^*$ with observational distributions $P$ and $P^*$, and $R$ a set of selection variables in $D$. The relation $Q = P^*(z|do(x),y)$ is transportable from $\Pi$ to $\Pi^*$ if and only if the expression $P(z|do(x),y,r)$ is reducible, using rules of do-calculus \cite{pearl} (Sec. 5.2), to an expression in which $R$ appears only as a conditioning variable in do-free terms. Given a DAG, open source tools like \cite{fusion} can check for causal effects transportability automatically.

As an illustration, consider Figure 6 (i). Let the variable $X$ denote artist and $Z$ denote the artwork. Suppose the variable $Y$ is an unobserved confounder denoting subjective emotions of the artist. Between any two artists, this variable $Y$ is bound to cause differences and hence the selection variable $R$ is pointing to $Y$ to indicate this difference. For this DAG, the causal effect of $X$ on $Z$ is not transportable or transferable across the two artists. For illustrative purposes, we will ignore the differences due to unobserved factors such as subjective emotions of artists and consider differences in observed variables. There can still be transportability bias as illustrated through the following case studies. 
\begin{figure}[t]
  \centering
    \includegraphics[width=0.38\textwidth]{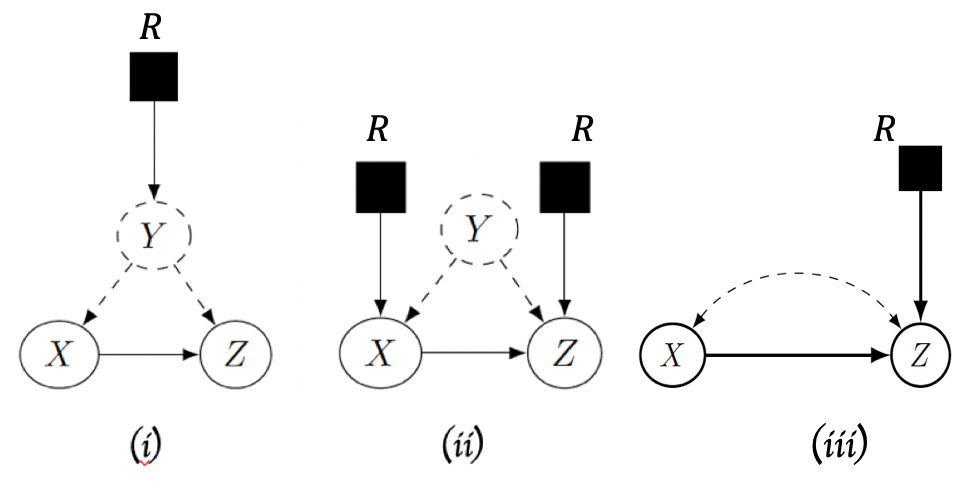}
    \caption{{\small (i): An example Selection diagram (SD). (ii): SD for case study 4 and 5. (iii) SD illustrating case study 10. Causal effect of $X$ on $Z$ is not identifiable across all scenarios (i), (ii), and (iii)}}
\end{figure}
\subsubsection{Case Study 4:} In \cite{cyclegan}, a model that can transfer photograph to an artist's style is proposed, say photograph and Cezanne for illustration. For simplicity, we consider only one genre say landscapes. The goal is to model Cezanne's style in rendering the landscape corresponding to the photo. Consider Figure 6 (ii) which illustrates one possible selection diagram for this case study. Let $X$ denote artist/photographer and let $Z$ denote the artwork/photo. Thus, there are two populations corresponding to the choice of $X$ and $Z$, i.e. photographer/photo, and artist/artwork. For the style transfer problem, we want to be able to capture the causal effect of the artist $X$ on the artwork $Z$ using the photograph. The shaded squares marked by the symbol $R$ are the selection variables and are used to denote differences in the two populations \cite{elias}. Further, there may other unobserved confounders. However, to illustrate biases beyond the difference in unobserved confounders, we will overlook such confounders in this case study.

The factors that influence an artist are different from those that influence a photographer. For example, Cezanne could be influenced by the art movement whereas the photographer may be influenced by the photography trends. This distinction is indicated by the selection variable $R$ pointing into $X$. Further, the factors that affect the artwork may be different in the two populations. A photograph may be subject to the camera characteristics, lighting, and measurement errors, selection variable $R$ pointing to $Z$ denotes this difference.  When the distinction is associated with the target variable, i.e. $Z$ in Figure 6 (ii), causal effects are not transportable \cite{aaai}. Thus, there will be transportability bias. 

In fact, in post-impressionism, the art movement primarily associated with Cezanne, artists had their own individual styles. As mentioned in \cite{james}, Cezanne concentrated on pictorial problems of creating depth in his landscapes. He used an organized system of layers to construct a series of horizontal planes, which build dimension and draw the viewer into the landscape. This technique is apparent in some of his works such as  {\it Mont Sainte-Victoire}, the {\it Viaduct of the Arc River Valley} and {\it The Gulf of Marseille Seen from L’Estaque} \cite{james}. In some of his works such as {\it Gardanne}, Cezanne painted the landscape with intense volumetric patterns of geometric rhythms most pronounced in the houses reflective of Cubism. The generated images in \cite{cyclegan} do not exhibit such geometric rhythms. 
\begin{figure}[t]
  \centering
    \includegraphics[width=0.45\textwidth]{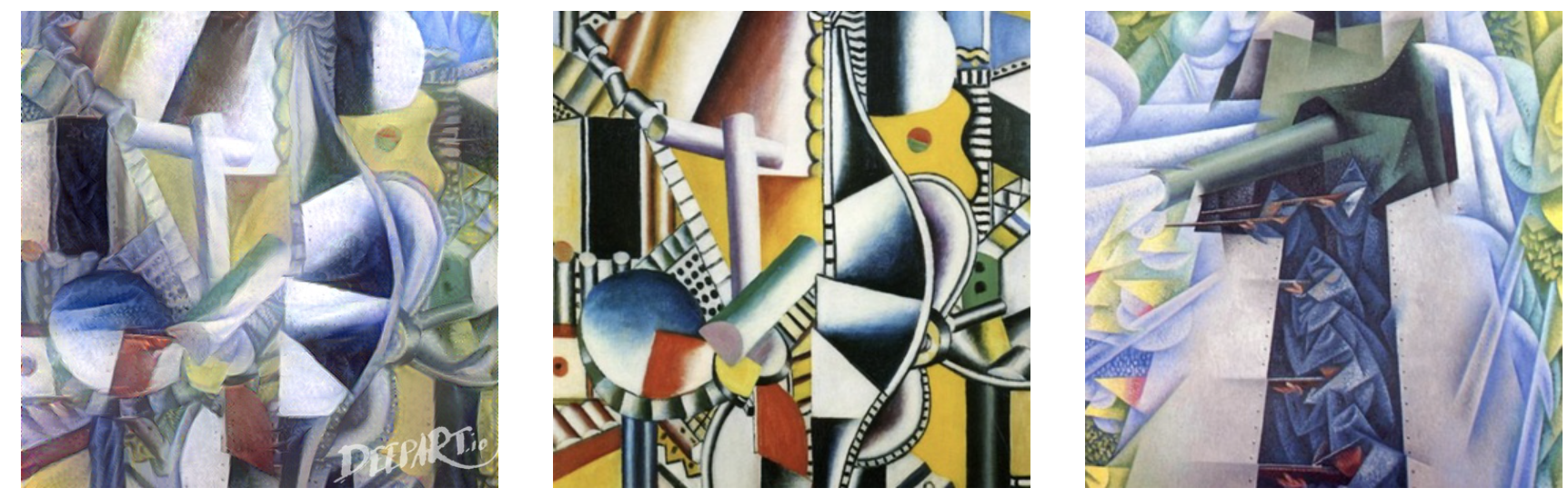}
    \caption{{\small Center: {\it ``Propellers"}, a Cubist work by Fernand Leger. Right: {\it ``Armoured Train in Action"}, a Futurism work by Gino Severini. Left: translation of the center image according to the style of right image by \cite{deepart}. Movement, a key aspect of Futurism, is missing in the translation. Image source: \cite{wikiart}}}
\end{figure}
\subsubsection{Case Study 5:} In the previous case study, we analyzed bias in the context of style transfer from one genre to another (from photograph to landscape). As another illustration, let us consider the problem of style transfer across art movements and genres. In order to demonstrate transportability bias in this setting, we consider DeepArt.io \cite{deepart}, an online platform that maps the style of one image to the content of the other using a neural network architecture \cite{gatys}. We consider a case study that involves Cubism and Futurism, two art movements in the modern art era. Both these movements had many common aspects, yet they diverged in subtle ways. Therefore, we find it to be an interesting case study for analysis. Both Cubism and Futurism focus on representation of objects from multiple perspectives/viewpoints and emphasize on geometrical shapes. Cubism, is concerned with forms in static relationships while Futurism is concerned with them in a kinetic state. Futurism emphasized on objects and events that involved movement such as wars, energy of nightclub, and so on \cite{imodern}. As such, we consider the following artworks for the case study.

{\it ``Propellers"} (center image in Figure 7) was a 1918 Cubist art by Fernand Leger. Leger was fascinated with technology, in particular with propellers, and  viewed them as objects of beauty holding them close to sculptures \cite{leger}. Right image in Figure 7 is a 1915 Futurism art by Gino Servini called {\it ``Armored Train in Action"}. Severini was inspired by Cubism but was a member of Futurism. Futurism used art as a symbol for expressing political and social views. Severini depicted aspects of war, movement, and modernity in this work. 

Figure 6 (ii) can be a potential DAG for this case study, note, there can be other DAGs based on assumptions. The differences in Cubism and Futurism art movements combined with the differences in genre influences the artists differently and the artwork differently. Also, there is the effect of unobserved factors such as the artist's emotions that influence the artist and the artwork. The left most image in Figure 7 corresponds to the ``Futurism version'' of {\it Propellers}.  Kinetic patterns which is a distinct feature of Futurism, is absent in the translated image. Given that the original image is that of a mechanical object (propellers), a Futurism version of it should have depicted the movement of the propellers much like in `Armoured train in Action' that shows the fractured landscape, which accentuates the train's force and momentum as it cuts through the countryside \cite{train}. Thus, there is bias in transferring styles.

\subsubsection{Case Study 6:} The previous case study encompassed style transfer across art movements which were similar in many ways. We now consider a case study involving style transfer between Realism and Expressionism, two art movements that have marked differences from one another. We consider common genre, namely portraits across the two art movements. Thus, this case study serves as a good test to see if style transfer from \cite{deepart, gatys} is effective given that the difference between two styles is significant.  
\begin{figure}[t]
  \centering
    \includegraphics[width=0.45\textwidth]{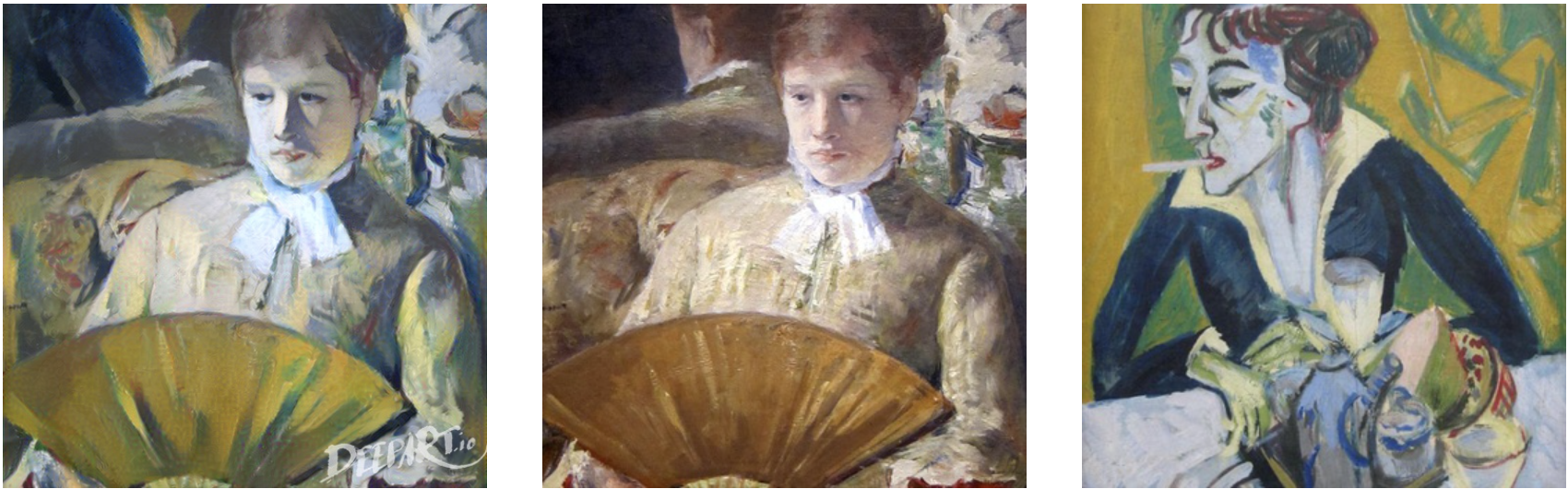}
    \caption{{\small Center: {\it ``Miss Mary Ellison"}, a Realism artwork by Mary Cassatt. Right: {\it ``Erna" }, an Expressionism artwork by Ernst Ludwig Kirchner. Left: translation of the center image according to the style of right image by \cite{deepart}. Distorted subjects, a key aspect of Expressionism is absent in the translation.  Image source: \cite{wikiart}}}
\end{figure}

Realism focuses on representing subject matter truthfully, without artificiality and avoiding implausible, exotic and supernatural elements \cite{wikiart}. The center image in Figure 8 is an Realism portrait by Mary Cassatt. Expressionists, on the other hand, used gestural brushstrokes and distorted subjects to portray intense emotions through their works. The right image in Figure 8 is an expressionism portrait by Ernst Ludwig Kirchner. As can be observed, the facial features in the right image have been distorted (e.g. sharp chin resulting in an almost triangular facial contour, pointed nose and ears) to intensify emotions. The left image in Figure 8 is the style translated version of the center image. Aesthetic innovations typical of an avante-garde movement like Expressionism are not evident in the style translated version.  As Kirchner himself said, in Expressionism, objective correctness of things is not emphasized \cite{story}, rather a new appearance is created through radical distortions of subjects to evoke intense emotional experiences. The style translated version is exactly as the original but for some color changes. The image neither exhibits any distorted features nor has gestural brushstrokes that portray intense emotions.  Thus, the style translation does not capture the subtleties of the Expressionism art movement.
\subsubsection{Case Study 7:}
As another case study to demonstrate biases in ``style" transfer, we consider ``GoART" \cite{goart}. This app allows a user to convert an uploaded photo into various styles spanning art movements such as Byzantine, Expressionism, Cubism, Ukiyo-e, and artists such as Van Gogh. We consider a 1970 folk art by Clementine Hunter titled {\it ``Black Matriarch"} shown in Figure 9 (i). Figure 9 (ii) shows the ``Expressionism" version of the {\it ``Black Matriarch"} from \cite{goart}. As can be noticed, the face is tinted in red. However, this kind of effect is not pronounced in light colored faces. Consider Figure 9 (iii). This is an early Renaissance sculpture by Desiderio da Settignano. The face in the ``Expressionism" version of this sculpture (Figure 9 (iv)) does not seem to have shades of red as significant as in Figure 9 (ii). Similar results were observed with other styles such as Byzantine wherein the face of {\it ``Black Matriarch"} was tinted with shades of blue while the fairer faces were not heavily tinted. There is noticeable difference in the way faces are converted across styles based on the color of the face, indicating potential racial biases.
\begin{figure}[t]
  \centering
    \includegraphics[width=0.45\textwidth]{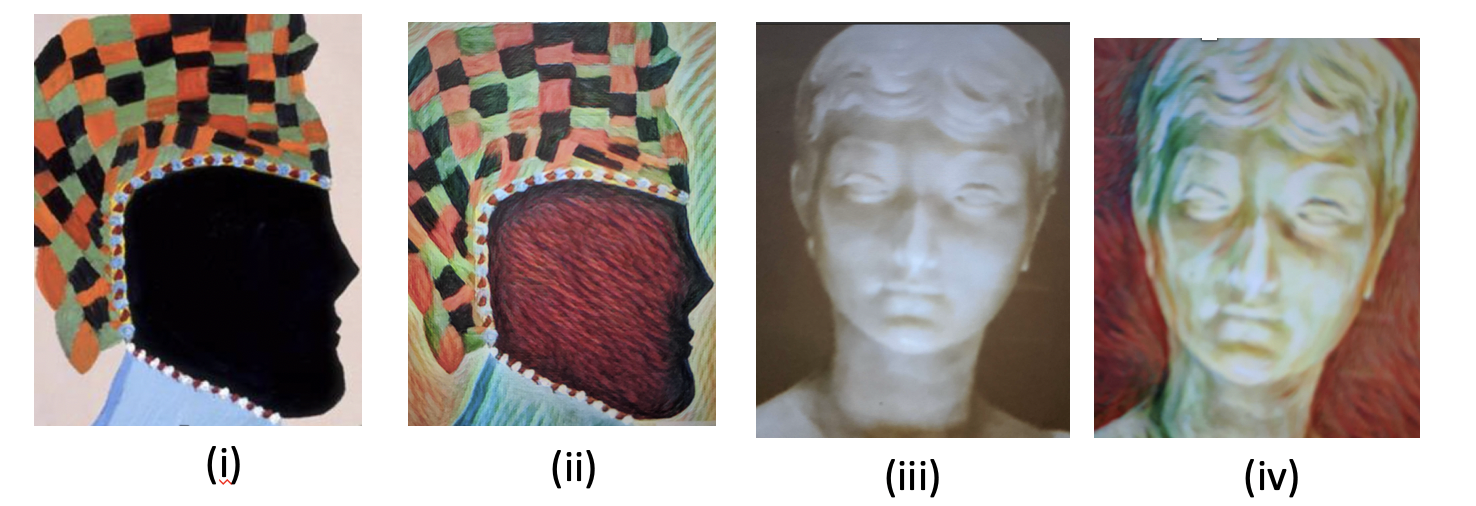}
    \caption{{\small (i): {\it ``Black Matriarch"}, a Folkart by Clementine Hunter. (ii): ``Expressionism version" of (i) by \cite{goart}. (iii): {\it ``Giovinetto"}, a Renaissance sculpture by Desiderio da Settignano. (iv): ``Expressionism version" of (iii) by \cite{goart}. Face color of {\it ``Black Matriarch"} is changed in the translation unlike in {\it ``Giovinetto"}, a white sculpture. Image source: \cite{wikiart}}}
\end{figure}
\subsubsection{Case Study 8:} We consider Abacus.AI's online tool that converts a user uploaded photo into a different gender: male to female and vice versa \cite{abacus}. One of their demos shows translation of the Renaissance painting of {\it Mona Lisa} into a masculine face. Thus, we experimented with other Renaissance paintings.  Figure 10 (i) is a `portrait of a man holding an apple' by Raphael and Figure 10 (iii) is a `portrait of a young man' by Italian artist Piero di Cosimo \cite{wikiart}. Figure 10 (ii) and 10 (iv) are the gender translated versions of (i) and (iii) by \cite{abacus}, both of which fail to identify the original paintings as those of men. Young men with long hair were mistaken to be women and thus the gender-translated versions of these images depicted masculine faces with beards. In the Renaissance era, it was common for young men to have long hair, often extending from ears to shoulders. Thus, by not understanding the differences due to culture across genders and age, men are being stereotyped as having short hair and \cite{abacus} thus exhibits transportability bias. 
\begin{figure}[h]
  \centering
    \includegraphics[width=0.45\textwidth]{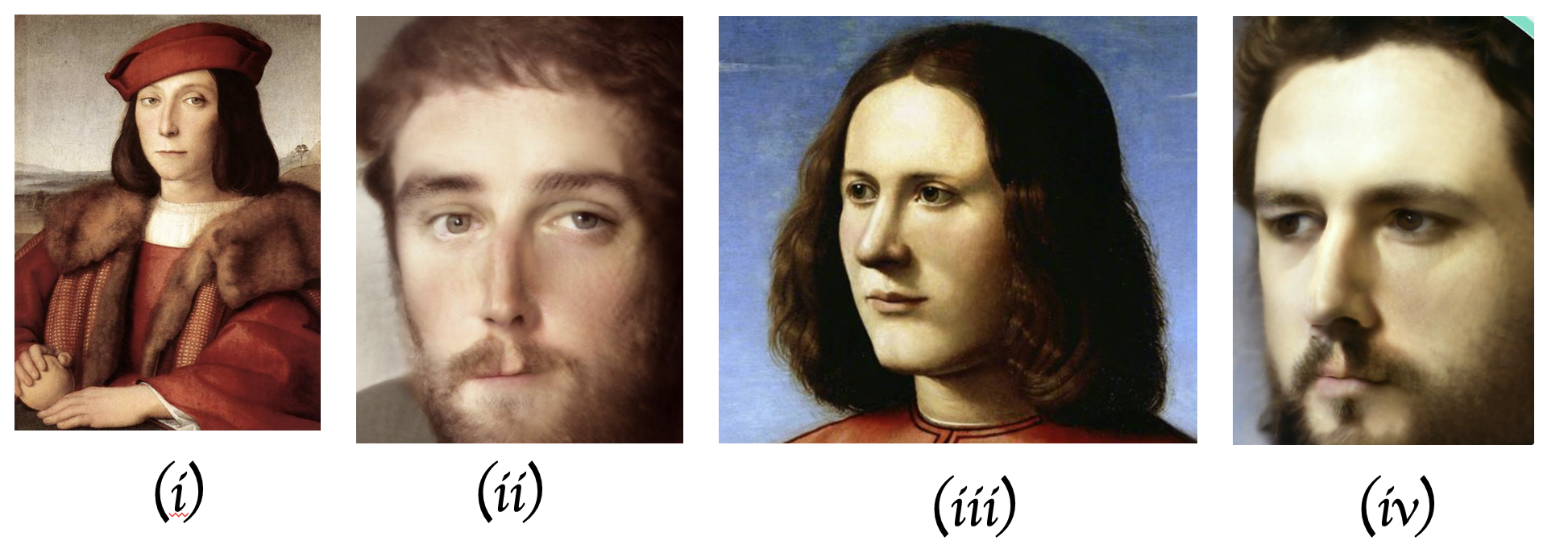}
    \caption{{\small (i) Portrait of a man by Raphael, (iii) Portrait of a young man by Cosimo. (ii), (iv): Gender translations of (i) and (iii) respectively. Young men with long hair were mistaken as women by \cite{abacus}} }
\end{figure}
\section{Biases in Datasets}
In this section, we discuss biases due to unrepresentative datasets and due to inconsistencies in annotation.
\subsection{Representational bias}
The bias that arises because of having a dataset that is not representative of the real world is referred to as representational bias. This is a particular type of selection bias. Specifically in the context of art datasets, there may be imbalances with respect to art genres (e.g. large number of photographs vs few sculptures), artists (e.g. mostly European artists vs few native artists), art movements (large number of works concerning Renaissance and modern art movements as opposed to others), and so on. The availability of artworks is one of the main constraints in collecting a dataset that is representative of the bygone times, but preferences of the dataset curators can also play a role in contributing to bias.

\subsubsection{Case Study 9:} Consider \cite{aiportraits} that was trained using about 45000 Renaissance portraits of mostly white people \cite{vice, sung}. Quite naturally, the system performs poorly on dark skinned people. Faces depicting different races, appearances, etc. have not been pooled into the dataset, thus contributing to representational bias. This is a particular instance of selection bias that has to do with dataset curation. 
Algorithms trained using datasets with severe class imbalances are bound to be biased. Figure 5 (iii) illustrates this. Suppose $X$ denotes artist and $Z$ denotes artworks, then class imbalances corresponds to $Z \rightarrow S$, i.e. type of artworks influences selection into the dataset (in this case mostly white Renaissance portraits were selected into the dataset). As condition (iii) fails in selection backdoor criteria, there is representational bias. 
\subsection{Label bias}
This type of bias is associated with the inconsistencies in the labeling process. Different annotators may have different preferences which can get reflected in the labels created. A common instance of label bias arises when different annotations could be used to represent an artwork. For example, a scene with clouds may be annotated as a ``cloudscape" by some annotators and more generally as a "landscape" by others.  

Yet another type of label bias arises when subjective biases of evaluators can affect labeling. ``Confirmation bias" \cite{plous}, a type of human bias, is closely related to this type of label bias. For example, in a task of annotating emotions associated with artworks, the labels could be biased by the subjective preferences of annotators such as their culture, beliefs, and introspective capabilities. Consider the Behance Artistic Media dataset \cite{behance} which provides labels based on emotions such as ``happy", ``scary", "peaceful", etc. Such labels could be based on annotator's beliefs, and can therefore be noisy. 

\subsubsection{Case Study 10:} To illustrate how annotation inconsistencies can induce bias, consider the ArtGAN model \cite{artgan}. This model uses the label information to train the GAN's discriminator. The authors claim that by using labels pertaining to genre (cityscapes, portraits, etc.), art media (e.g. sketch and study, engraving), and style (e.g. Ukiyo-e), they are able to generate images of those categories.  However, the annotations are not necessarily reliable indicators of genres or styles. For example, ``sketch and study" category includes several other categories such as portraits, religious paintings, and allegorical works to name a few. Figure 6 (iii) illustrates this bias. Let  $\Pi$ and $\Pi^*$ denote the environments corresponding to two different annotators. Suppose $X$ denotes art movement's style (e.g. Ukiyo-e), and $Z$ is the artwork. Annotation inconsistencies across annotators affects the artworks' labels, this is indicated by the selection variable $R$ pointing to $Z$. 
When there are differences in the target variable (i.e. label of artworks), the causal effect of $X$ on $Z$ is not identifiable using the annotations provided \cite{aaai}. Thus, a generative model that leverages such labels in modeling style is prone to bias.
\section{Discussion}
Art is much more than an aesthetic entity. As elaborated in \cite{young}, art imparts ``moral knowledge", i.e. knowledge about what ought to do and not to do \cite{tim}. Art also enables ``emphathic knowledge", something through which one can compare different views of the world through direct experience \cite{novitz}. Art initiates a conversation with the public \cite{aies}, it is a form of language that is not just a mimicry but a symbolic transposition \cite{leroi}.  Art is {\it `a form of technology that contributes to knowledge production by exemplifying aspects of the world that would otherwise go overlooked'} \cite{tim}. Thus given its powerful impact in shaping moral and empathetic values, generative art comes with the responsibility of creating art that respects and upholds societal ethics. By coloring the face of the ``Black Matriarch", \cite{goart} is not only depicting racial bias, but also inaccurate in its representation of the art movement. As artist Edgar Degas remarks, {\it ``Art is  not just what you see but what you make others see"}. Thus, generative art that does not promote diversity and inclusiveness has the potential of creating and communicating unethical values. 

Art reflects cognitive abilities of the artist \cite{dissan}. Cognitive abilities  include perception, memory, emotions, and other latent aspects about the artist.  Generative art that is meant to create art in the ``style" of various artists must reflect and respect artist's cognitive abilities and not stereotype them based on any narrowly defined metric. For example, often, ``style of Van Gogh" is largely modeled based on the brushstrokes in his rendition {\it ``Starry Nights"} or based on certain colors such as in \cite{artgan}. Similarly, ``style of Cezanne" in \cite{cyclegan} little reflects the variety of geometric patterns that were prominent in his works. Needless to mention that cognitive aspects of the artist are not considered. As discussed in Section 4 and 6, majority of these issues arise due to framing effect bias and algorithmic bias. Often style is defined and modeled in a way to suit the algorithm's performance. Not only are several important abilities and achievements of artists overlooked in the process of poor style modeling, but also those pertaining to larger art movements are ignored. For example, advanced techniques such as exaggerated foreshortening and perspective modeling that were typical of many Ukiyo-e renditions are hardly visible in the generated versions \cite{cyclegan, goart}. 

In a recent photo booth titled {\it ``Latent Face"}, latent vectors of the styleGAN model were combined to generate "hypothetical children" of subjects depicted in the original portraits \cite{stylegan}. While this may have been an exploratory experiment, the task exemplifies framing effect bias. Defining ``children" as combination of latent vectors is highly questionable. The latent vectors may or may not have any reliable interpretation, and a very difficult and potentially impossible problem of generating faces of hypothetical children is conveniently framed as a simple task. There are also several ethical concerns associated with such a framing given that people depicted in the original portraits were not related or never had children.  

Framing effect bias coupled with algorithmic bias contributes to inaccurate knowledge about history. Artworks were often meant to document important events in history such as wars, mythological events, political movements, etc.  By wrongly modeling or overlooking certain subtle aspects, generative art can contribute to false perceptions about social, cultural and political aspects of past times and hinder awareness about important historical events. For example, as discussed in case study 5, in the Futurism artwork {\it ``Armored Train in Action"}, artist Severini conveys his views on war that was prevalent during the time. In fact, Futurism artists heavily depicted their views of political events through patterns indicating movement in the artworks. A generated style translation should thus preserve such important characteristics of art movements, or else they will be contributing to a bias in understanding history. People have a propensity to favor suggestions from automated systems and to ignore contradictory information, even if it is correct. This is called as {\it ``automation bias"} \cite{automation}. Thus, people may give little importance to historical evidence. Further, often, evaluation of generative art is done by people (e.g. Amazon Mechanical Turk workers) who do not necessarily possess domain knowledge. Thus, even if generated art is not accurately representing cultural and historical knowledge, people may endorse them.

Tutorials like \cite{timnit} and \cite{moss} emphasize on the need to inspect the design choices made by the creators of AI systems and the socio-political contexts that shape their deployment. Can algorithms accurately model artist's ``styles"? More broadly, is the defined problem even solvable? How can representative datasets and reliable labels be created to address the defined problem? What are the measurement biases in digitally capturing art? What are the socio-cultural impacts of generative art? Does generative art promote inclusiveness? Who should own responsibility for biases in generative art? These are just some questions that need to be analyzed. Also, domain experts (e.g. art historians) should be involved in the generative art pipeline to better inform the process of art creation.
\vspace{-0.05in}
\section{Conclusions}
In this paper, we investigated biases in the generative art AI pipeline from the perspective of art history. Leveraging structural causal models, we highlighted how current methods fall short in modeling the process of art creation and illustrated instances of framing effect bias, dataset bias, selection bias, confounding bias, and transportability bias.  We also discussed the socio-cultural impacts of these biases. We hope our work sparks inter-disciplinary dicussions and inspires new directions concerning accountability of generative art.

\bibliographystyle{ACM-Reference-Format}
\bibliography{sample-base}


\begin{thebibliography}{70}


\ifx \showCODEN    \undefined \def \showCODEN     #1{\unskip}     \fi
\ifx \showDOI      \undefined \def \showDOI       #1{#1}\fi
\ifx \showISBNx    \undefined \def \showISBNx     #1{\unskip}     \fi
\ifx \showISBNxiii \undefined \def \showISBNxiii  #1{\unskip}     \fi
\ifx \showISSN     \undefined \def \showISSN      #1{\unskip}     \fi
\ifx \showLCCN     \undefined \def \showLCCN      #1{\unskip}     \fi
\ifx \shownote     \undefined \def \shownote      #1{#1}          \fi
\ifx \showarticletitle \undefined \def \showarticletitle #1{#1}   \fi
\ifx \showURL      \undefined \def \showURL       {\relax}        \fi
\providecommand\bibfield[2]{#2}
\providecommand\bibinfo[2]{#2}
\providecommand\natexlab[1]{#1}
\providecommand\showeprint[2][]{arXiv:#2}

\bibitem[\protect\citeauthoryear{Abacus.AI}{Abacus.AI}{2020}]%
        {abacus}
\bibfield{author}{\bibinfo{person}{Abacus.AI}.}
  \bibinfo{year}{2020}\natexlab{}.
\newblock \showarticletitle{Effortlessly embed cutting edge AI in your
  Applications}. In \bibinfo{booktitle}{\emph{https://abacus.ai}}.
\newblock


\bibitem[\protect\citeauthoryear{AIportraits}{AIportraits}{2020}]%
        {aiportraits}
\bibfield{author}{\bibinfo{person}{AIportraits}.}
  \bibinfo{year}{2020}\natexlab{}.
\newblock \showarticletitle{AIportraits: The easiest way to make your portraits
  look stunning}.
\newblock \bibinfo{journal}{\emph{https://aiportraits.org}}
  (\bibinfo{year}{2020}).
\newblock


\bibitem[\protect\citeauthoryear{Artbreeder}{Artbreeder}{2020}]%
        {artbreeder}
\bibfield{author}{\bibinfo{person}{Artbreeder}.}
  \bibinfo{year}{2020}\natexlab{}.
\newblock \showarticletitle{Artbreeder: Extend your Imagination.}
\newblock \bibinfo{journal}{\emph{https://www.artbreeder.com}}
  (\bibinfo{year}{2020}).
\newblock


\bibitem[\protect\citeauthoryear{Asendorf}{Asendorf}{1994}]%
        {leger}
\bibfield{author}{\bibinfo{person}{Christoph Asendorf}.}
  \bibinfo{year}{1994}\natexlab{}.
\newblock \showarticletitle{The Propeller and the Avant-Garde: Leger, Duchamp,
  Brancusi}.
\newblock \bibinfo{journal}{\emph{Fernand leger: The Rhythm of Modern Life}}
  (\bibinfo{year}{1994}).
\newblock


\bibitem[\protect\citeauthoryear{Bailey}{Bailey}{2019}]%
        {stylegan}
\bibfield{author}{\bibinfo{person}{Jason Bailey}.}
  \bibinfo{year}{2019}\natexlab{}.
\newblock \showarticletitle{Breeding Paintings With Machine Learning}.
\newblock
  \bibinfo{journal}{\emph{https://www.artnome.com/news/2019/8/25/breeding-paintings-with-machine-learning}}
  (\bibinfo{year}{2019}).
\newblock


\bibitem[\protect\citeauthoryear{Bailey}{Bailey}{2020a}]%
        {predict}
\bibfield{author}{\bibinfo{person}{Jason Bailey}.}
  \bibinfo{year}{2020}\natexlab{a}.
\newblock \showarticletitle{2020 Art Market Predictions}.
\newblock
  \bibinfo{journal}{\emph{https://www.artnome.com/news/2020/1/27/2020-art-market-predictions}}
  (\bibinfo{year}{2020}).
\newblock


\bibitem[\protect\citeauthoryear{Bailey}{Bailey}{2020b}]%
        {history}
\bibfield{author}{\bibinfo{person}{Jason Bailey}.}
  \bibinfo{year}{2020}\natexlab{b}.
\newblock \showarticletitle{The tools of generative art from flash to neural
  networks}.
\newblock
  \bibinfo{journal}{\emph{https://www.artnews.com/art-in-america/features/generative-art-tools-flash-processing-neural-networks-1202674657/}}
  (\bibinfo{year}{2020}).
\newblock


\bibitem[\protect\citeauthoryear{Bareinboim et~al\mbox{.}}{Bareinboim
  et~al\mbox{.}}{2020}]%
        {fusion}
\bibfield{author}{\bibinfo{person}{Elias Bareinboim} {et~al\mbox{.}}}
  \bibinfo{year}{2020}\natexlab{}.
\newblock \showarticletitle{Causal Fusion}. In
  \bibinfo{booktitle}{\emph{https://causalfusion.net}}.
\newblock


\bibitem[\protect\citeauthoryear{Bareinboim and Pearl}{Bareinboim and
  Pearl}{2012}]%
        {aaai}
\bibfield{author}{\bibinfo{person}{Elias Bareinboim} {and}
  \bibinfo{person}{Judea Pearl}.} \bibinfo{year}{2012}\natexlab{}.
\newblock \showarticletitle{Transportability of Causal Effects: Completeness
  Results}. In \bibinfo{booktitle}{\emph{AAAI}}.
\newblock


\bibitem[\protect\citeauthoryear{Bareinboim and Pearl}{Bareinboim and
  Pearl}{2016}]%
        {elias}
\bibfield{author}{\bibinfo{person}{Elias Bareinboim} {and}
  \bibinfo{person}{Judea Pearl}.} \bibinfo{year}{2016}\natexlab{}.
\newblock \showarticletitle{Causal inference and the data-fusion problem}.
\newblock \bibinfo{journal}{\emph{Proceedings of the National Academy of
  Sciences}} (\bibinfo{year}{2016}).
\newblock


\bibitem[\protect\citeauthoryear{Boden and Edmonds}{Boden and Edmonds}{2009}]%
        {genart}
\bibfield{author}{\bibinfo{person}{Margaret Boden} {and}
  \bibinfo{person}{Ernest Edmonds}.} \bibinfo{year}{2009}\natexlab{}.
\newblock \showarticletitle{What is Generative Art}.
\newblock \bibinfo{journal}{\emph{Digital Creativity}} (\bibinfo{year}{2009}).
\newblock


\bibitem[\protect\citeauthoryear{Brooks}{Brooks}{2017}]%
        {mit}
\bibfield{author}{\bibinfo{person}{Rodney Brooks}.}
  \bibinfo{year}{2017}\natexlab{}.
\newblock \showarticletitle{The Seven Deadly Sins of AI Predictions}.
\newblock \bibinfo{journal}{\emph{MIT Technology Review}}
  (\bibinfo{year}{2017}).
\newblock


\bibitem[\protect\citeauthoryear{C}{C}{2019}]%
        {market}
\bibfield{author}{\bibinfo{person}{McAndrew C}.}
  \bibinfo{year}{2019}\natexlab{}.
\newblock \showarticletitle{The art market. An Art Basel and UBS Report}.
\newblock
  \bibinfo{journal}{\emph{https://www.artbasel.com/about/initiatives/the-art-market}}
  (\bibinfo{year}{2019}).
\newblock


\bibitem[\protect\citeauthoryear{C and B}{C and B}{2007}]%
        {reas}
\bibfield{author}{\bibinfo{person}{Reas C} {and} \bibinfo{person}{Fry B}.}
  \bibinfo{year}{2007}\natexlab{}.
\newblock \showarticletitle{Processing: a programming handbook for visual
  designers and artists}.
\newblock \bibinfo{journal}{\emph{MIT Press}} (\bibinfo{year}{2007}).
\newblock


\bibitem[\protect\citeauthoryear{Christies}{Christies}{2018}]%
        {christies}
\bibfield{author}{\bibinfo{person}{Christies}.}
  \bibinfo{year}{2018}\natexlab{}.
\newblock \showarticletitle{Is artificial intelligence set to become art's next
  medium?}
\newblock \bibinfo{journal}{\emph{https://goo.gl/4LDZjX}}
  (\bibinfo{year}{2018}).
\newblock


\bibitem[\protect\citeauthoryear{Coeckelbergh}{Coeckelbergh}{2017}]%
        {mark}
\bibfield{author}{\bibinfo{person}{Mark Coeckelbergh}.}
  \bibinfo{year}{2017}\natexlab{}.
\newblock \showarticletitle{Can Machines Create Art?}
\newblock \bibinfo{journal}{\emph{Philosophy and Technology}}
  (\bibinfo{year}{2017}).
\newblock


\bibitem[\protect\citeauthoryear{Crawford and Paglen}{Crawford and
  Paglen}{2019}]%
        {kate}
\bibfield{author}{\bibinfo{person}{Kate Crawford} {and} \bibinfo{person}{Trevor
  Paglen}.} \bibinfo{year}{2019}\natexlab{}.
\newblock \showarticletitle{Excavating AI: The Politics of Images in Machine
  Learning Training Sets}.
\newblock \bibinfo{journal}{\emph{https://www.excavating.ai}}
  (\bibinfo{year}{2019}).
\newblock


\bibitem[\protect\citeauthoryear{D}{D}{1987}]%
        {novitz}
\bibfield{author}{\bibinfo{person}{Novitz D}.} \bibinfo{year}{1987}\natexlab{}.
\newblock \showarticletitle{Knowledge, Fiction, and Imagination}.
\newblock \bibinfo{journal}{\emph{Temple University Press}}
  (\bibinfo{year}{1987}).
\newblock


\bibitem[\protect\citeauthoryear{Daniele and Song}{Daniele and Song}{2019}]%
        {aies}
\bibfield{author}{\bibinfo{person}{Antonio Daniele} {and}
  \bibinfo{person}{Yi-Zhe Song}.} \bibinfo{year}{2019}\natexlab{}.
\newblock \showarticletitle{AI+Art= Human}.
\newblock \bibinfo{journal}{\emph{AAAI AI Ethics and Society}}
  (\bibinfo{year}{2019}).
\newblock


\bibitem[\protect\citeauthoryear{Deepart.io}{Deepart.io}{2020}]%
        {deepart}
\bibfield{author}{\bibinfo{person}{Deepart.io}.}
  \bibinfo{year}{2020}\natexlab{}.
\newblock \showarticletitle{Deepart.io}.
\newblock \bibinfo{journal}{\emph{https://deepart.io}} (\bibinfo{year}{2020}).
\newblock


\bibitem[\protect\citeauthoryear{Dissanayake}{Dissanayake}{2001}]%
        {dissan}
\bibfield{author}{\bibinfo{person}{E. Dissanayake}.}
  \bibinfo{year}{2001}\natexlab{}.
\newblock \showarticletitle{Where art comes from and Why}.
\newblock \bibinfo{journal}{\emph{University of Washington Press}}
  (\bibinfo{year}{2001}).
\newblock


\bibitem[\protect\citeauthoryear{Dunbabin}{Dunbabin}{1999}]%
        {mosaic}
\bibfield{author}{\bibinfo{person}{Katherine Dunbabin}.}
  \bibinfo{year}{1999}\natexlab{}.
\newblock \showarticletitle{Mosaics of the Greek and Roman World}.
\newblock \bibinfo{journal}{\emph{Cambridge University Press}}
  (\bibinfo{year}{1999}).
\newblock


\bibitem[\protect\citeauthoryear{(Editor)}{(Editor)}{2007}]%
        {canon}
\bibfield{author}{\bibinfo{person}{Anna~Brzyski (Editor)}.}
  \bibinfo{year}{2007}\natexlab{}.
\newblock \showarticletitle{Partisan Canons}.
\newblock \bibinfo{journal}{\emph{Duke University Press}}
  (\bibinfo{year}{2007}).
\newblock


\bibitem[\protect\citeauthoryear{Elgammal}{Elgammal}{2019}]%
        {elgammalnyc}
\bibfield{author}{\bibinfo{person}{Ahmed Elgammal}.}
  \bibinfo{year}{2019}\natexlab{}.
\newblock \showarticletitle{Faceless Portraits Transcending Time}.
\newblock \bibinfo{journal}{\emph{HG Contemporary New York
  https://uploads.strikinglycdn.com/files/3e2cdfa0-8b8f-44ea-a6ca-d12f123e3b0c/AICAN-HG-Catalogue-web.pdf}}
  (\bibinfo{year}{2019}).
\newblock


\bibitem[\protect\citeauthoryear{Elgammal, Liu, Elhoseiny, and
  Mazzone}{Elgammal et~al\mbox{.}}{2017}]%
        {mazzone}
\bibfield{author}{\bibinfo{person}{Ahmed Elgammal}, \bibinfo{person}{Bingchen
  Liu}, \bibinfo{person}{Mohamed Elhoseiny}, {and} \bibinfo{person}{Marian
  Mazzone}.} \bibinfo{year}{2017}\natexlab{}.
\newblock \showarticletitle{CAN: Creative Adversarial Networks, Generating
  "Art" by Learning About Styles and Deviating from Style Norms}.
\newblock \bibinfo{journal}{\emph{International Conference on Computational
  Creativity (ICCC)}} (\bibinfo{year}{2017}).
\newblock


\bibitem[\protect\citeauthoryear{Gatys, Ecker, and Bethge}{Gatys
  et~al\mbox{.}}{2016}]%
        {gatys}
\bibfield{author}{\bibinfo{person}{Leon~A. Gatys}, \bibinfo{person}{Alexander~S
  Ecker}, {and} \bibinfo{person}{Matthias Bethge}.}
  \bibinfo{year}{2016}\natexlab{}.
\newblock \showarticletitle{Image Style Transfer Using Convolutional Neural
  Networks.}
\newblock \bibinfo{journal}{\emph{Computer Vision and Pattern Recognition}}
  (\bibinfo{year}{2016}).
\newblock


\bibitem[\protect\citeauthoryear{Gebru and Denton}{Gebru and Denton}{2020}]%
        {timnit}
\bibfield{author}{\bibinfo{person}{Timnit Gebru} {and} \bibinfo{person}{Emily
  Denton}.} \bibinfo{year}{2020}\natexlab{}.
\newblock \showarticletitle{Tutorial on Fairness Accountability Transparency
  and Ethics in Computer Vision}.
\newblock \bibinfo{journal}{\emph{CVPR Tutorial}} (\bibinfo{year}{2020}).
\newblock


\bibitem[\protect\citeauthoryear{Glymour and Herington}{Glymour and
  Herington}{2019}]%
        {glymour}
\bibfield{author}{\bibinfo{person}{Bruce Glymour} {and}
  \bibinfo{person}{Jonathan Herington}.} \bibinfo{year}{2019}\natexlab{}.
\newblock \showarticletitle{Measuring the Biases that Matter}.
\newblock \bibinfo{journal}{\emph{FaccT}} (\bibinfo{year}{2019}).
\newblock


\bibitem[\protect\citeauthoryear{GoART}{GoART}{2020}]%
        {goart}
\bibfield{author}{\bibinfo{person}{GoART}.} \bibinfo{year}{2020}\natexlab{}.
\newblock \showarticletitle{GoART: AI Photo Effects}.
\newblock
  \bibinfo{journal}{\emph{http://goart.fotor.com.s3-website-us-west-2.amazonaws.com}}
  (\bibinfo{year}{2020}).
\newblock


\bibitem[\protect\citeauthoryear{Goodfellow, Pouget-Abadie, Mirza, Xu,
  Warde-Farley, Ozair, Courville, and Bengio}{Goodfellow et~al\mbox{.}}{2014}]%
        {goodfellow}
\bibfield{author}{\bibinfo{person}{Ian~J. Goodfellow}, \bibinfo{person}{Jean
  Pouget-Abadie}, \bibinfo{person}{Mehdi Mirza}, \bibinfo{person}{Bing Xu},
  \bibinfo{person}{David Warde-Farley}, \bibinfo{person}{Sherjil Ozair},
  \bibinfo{person}{Aaron Courville}, {and} \bibinfo{person}{Yoshua Bengio}.}
  \bibinfo{year}{2014}\natexlab{}.
\newblock \showarticletitle{Generative adversarial nets}.
\newblock \bibinfo{journal}{\emph{NeurIPS}} (\bibinfo{year}{2014}).
\newblock


\bibitem[\protect\citeauthoryear{Gorichanaz}{Gorichanaz}{2020}]%
        {tim}
\bibfield{author}{\bibinfo{person}{Tim Gorichanaz}.}
  \bibinfo{year}{2020}\natexlab{}.
\newblock \showarticletitle{Engaging with Public Art: An Exploration of the
  Design Space}.
\newblock \bibinfo{journal}{\emph{CHI}} (\bibinfo{year}{2020}).
\newblock


\bibitem[\protect\citeauthoryear{Ha and Eck}{Ha and Eck}{2018}]%
        {magenta}
\bibfield{author}{\bibinfo{person}{David Ha} {and} \bibinfo{person}{Douglas
  Eck}.} \bibinfo{year}{2018}\natexlab{}.
\newblock \showarticletitle{A Neural Representation of Sketch Drawings}.
\newblock \bibinfo{journal}{\emph{International Conference on Learning
  Representations}} (\bibinfo{year}{2018}).
\newblock


\bibitem[\protect\citeauthoryear{Haeberli}{Haeberli}{1990}]%
        {paul}
\bibfield{author}{\bibinfo{person}{Paul Haeberli}.}
  \bibinfo{year}{1990}\natexlab{}.
\newblock \showarticletitle{Paint By Numbers: Abstract Image Representations}.
\newblock \bibinfo{journal}{\emph{SIGGRAPH}} (\bibinfo{year}{1990}).
\newblock


\bibitem[\protect\citeauthoryear{Hertzmann}{Hertzmann}{2018}]%
        {hertzman}
\bibfield{author}{\bibinfo{person}{Aaron Hertzmann}.}
  \bibinfo{year}{2018}\natexlab{}.
\newblock \showarticletitle{Can Computers create Art?}
\newblock \bibinfo{journal}{\emph{Arts}} (\bibinfo{year}{2018}).
\newblock


\bibitem[\protect\citeauthoryear{Hertzmann, Jacobs, Oliver, Curless, and
  Salesin}{Hertzmann et~al\mbox{.}}{2001}]%
        {jacobs}
\bibfield{author}{\bibinfo{person}{Aaron Hertzmann}, \bibinfo{person}{C.
  Jacobs}, \bibinfo{person}{N. Oliver}, \bibinfo{person}{B. Curless}, {and}
  \bibinfo{person}{D.H. Salesin}.} \bibinfo{year}{2001}\natexlab{}.
\newblock \showarticletitle{Image Analogies}.
\newblock \bibinfo{journal}{\emph{SIGGRAPH}} (\bibinfo{year}{2001}).
\newblock


\bibitem[\protect\citeauthoryear{Hong}{Hong}{2018}]%
        {hong}
\bibfield{author}{\bibinfo{person}{Joo-Wha Hong}.}
  \bibinfo{year}{2018}\natexlab{}.
\newblock \showarticletitle{Bias in Perception of Art Produced by Artificial
  Intelligence}.
\newblock \bibinfo{journal}{\emph{International Conference on Human Computer
  Interaction}} (\bibinfo{year}{2018}).
\newblock


\bibitem[\protect\citeauthoryear{Instapainting}{Instapainting}{2020}]%
        {aipainter}
\bibfield{author}{\bibinfo{person}{Instapainting}.}
  \bibinfo{year}{2020}\natexlab{}.
\newblock \showarticletitle{AI painter}.
\newblock \bibinfo{journal}{\emph{https://www.instapainting.com/ai-painter}}
  (\bibinfo{year}{2020}).
\newblock


\bibitem[\protect\citeauthoryear{Jr}{Jr}{2019}]%
        {vice}
\bibfield{author}{\bibinfo{person}{Edward~Ongweso Jr}.}
  \bibinfo{year}{2019}\natexlab{}.
\newblock \showarticletitle{Racial Bias in AI Isn’t Getting Better and
  Neither Are Researchers’ Excuses}.
\newblock
  \bibinfo{journal}{\emph{https://www.vice.com/en\_us/article/8xzwgx/racial-bias-in-ai-isnt-getting-better-and-neither-are-researchers-excuses}}
  (\bibinfo{year}{2019}).
\newblock


\bibitem[\protect\citeauthoryear{Kaeser-Chen, Dubois, Schuur, and
  Moss}{Kaeser-Chen et~al\mbox{.}}{2020}]%
        {moss}
\bibfield{author}{\bibinfo{person}{Christine Kaeser-Chen},
  \bibinfo{person}{Elizabeth Dubois}, \bibinfo{person}{Friederike Schuur},
  {and} \bibinfo{person}{Emanuel Moss}.} \bibinfo{year}{2020}\natexlab{}.
\newblock \showarticletitle{Positionality-aware machine learning: translation
  tutorial}.
\newblock \bibinfo{journal}{\emph{FAccT Tutorial}} (\bibinfo{year}{2020}).
\newblock


\bibitem[\protect\citeauthoryear{Kusner, Loftus, Russell, and Silva}{Kusner
  et~al\mbox{.}}{2017}]%
        {silva}
\bibfield{author}{\bibinfo{person}{Matt~J. Kusner}, \bibinfo{person}{Joshua~R.
  Loftus}, \bibinfo{person}{Chris Russell}, {and} \bibinfo{person}{Ricardo
  Silva}.} \bibinfo{year}{2017}\natexlab{}.
\newblock \showarticletitle{Counterfactual Fairness}.
\newblock \bibinfo{journal}{\emph{NeurIPS}} (\bibinfo{year}{2017}).
\newblock


\bibitem[\protect\citeauthoryear{Leroi-Gourhan}{Leroi-Gourhan}{1993}]%
        {leroi}
\bibfield{author}{\bibinfo{person}{A. Leroi-Gourhan}.}
  \bibinfo{year}{1993}\natexlab{}.
\newblock \showarticletitle{Gesture and Speech}.
\newblock \bibinfo{journal}{\emph{MIT Press}} (\bibinfo{year}{1993}).
\newblock


\bibitem[\protect\citeauthoryear{M}{M}{1977}]%
        {heidegger}
\bibfield{author}{\bibinfo{person}{Heidegger M}.}
  \bibinfo{year}{1977}\natexlab{}.
\newblock \showarticletitle{The question concerning technology, and other
  essays}.
\newblock \bibinfo{journal}{\emph{Garland Publishing INC}}
  (\bibinfo{year}{1977}).
\newblock


\bibitem[\protect\citeauthoryear{Macnish}{Macnish}{2020}]%
        {drawthis}
\bibfield{author}{\bibinfo{person}{Dan Macnish}.}
  \bibinfo{year}{2020}\natexlab{}.
\newblock \showarticletitle{Draw this}.
\newblock \bibinfo{journal}{\emph{https://danmacnish.com/drawthis/}}
  (\bibinfo{year}{2020}).
\newblock


\bibitem[\protect\citeauthoryear{Miller}{Miller}{2019}]%
        {miller}
\bibfield{author}{\bibinfo{person}{Authur Miller}.}
  \bibinfo{year}{2019}\natexlab{}.
\newblock \showarticletitle{The Artist in the Machine The World of AI-Powered
  Creativity}.
\newblock \bibinfo{journal}{\emph{MIT Press}} (\bibinfo{year}{2019}).
\newblock


\bibitem[\protect\citeauthoryear{Mordvintsev, Olah, and Tyka}{Mordvintsev
  et~al\mbox{.}}{2015}]%
        {deepdream}
\bibfield{author}{\bibinfo{person}{Alexander Mordvintsev},
  \bibinfo{person}{Christopher Olah}, {and} \bibinfo{person}{Mike Tyka}.}
  \bibinfo{year}{2015}\natexlab{}.
\newblock \showarticletitle{Deep Dream}.
\newblock \bibinfo{journal}{\emph{https://github.com/google/deepdream}}
  (\bibinfo{year}{2015}).
\newblock


\bibitem[\protect\citeauthoryear{of~Modern~Art}{of~Modern~Art}{2020}]%
        {train}
\bibfield{author}{\bibinfo{person}{MoMA:~Musuem of Modern~Art}.}
  \bibinfo{year}{2020}\natexlab{}.
\newblock \showarticletitle{Gino Severini: Armoured Train in Action}. In
  \bibinfo{booktitle}{\emph{https://www.moma.org/collection/works/33837}}.
\newblock


\bibitem[\protect\citeauthoryear{Online}{Online}{2020}]%
        {oxford}
\bibfield{author}{\bibinfo{person}{Oxford~Art Online}.}
  \bibinfo{year}{2020}\natexlab{}.
\newblock \showarticletitle{Impressionism and Post-Impressionism}.
\newblock
  \bibinfo{journal}{\emph{https://www.oxfordartonline.com/page/impressionism-and-post-impressionism/impressionism-and-postimpressionism}}
  (\bibinfo{year}{2020}).
\newblock


\bibitem[\protect\citeauthoryear{Pearl}{Pearl}{2009}]%
        {pearl}
\bibfield{author}{\bibinfo{person}{Judea Pearl}.}
  \bibinfo{year}{2009}\natexlab{}.
\newblock \showarticletitle{Causality: Models, Reasoning and Inference, 2nd
  Edition}.
\newblock \bibinfo{journal}{\emph{Cambridge University Press}}
  (\bibinfo{year}{2009}).
\newblock


\bibitem[\protect\citeauthoryear{Pearl and Bareinboim}{Pearl and
  Bareinboim}{2014}]%
        {transport}
\bibfield{author}{\bibinfo{person}{Judea Pearl} {and} \bibinfo{person}{Elias
  Bareinboim}.} \bibinfo{year}{2014}\natexlab{}.
\newblock \showarticletitle{External Validity: From Do-Calculus to
  Transportability Across Populations}.
\newblock \bibinfo{journal}{\emph{Statistical Sciences}}
  (\bibinfo{year}{2014}).
\newblock


\bibitem[\protect\citeauthoryear{Plous}{Plous}{1993}]%
        {plous}
\bibfield{author}{\bibinfo{person}{Scott Plous}.}
  \bibinfo{year}{1993}\natexlab{}.
\newblock \showarticletitle{The psychology of judgment and decision making.}
\newblock \bibinfo{journal}{\emph{Mc-Graw Hill}} (\bibinfo{year}{1993}).
\newblock


\bibitem[\protect\citeauthoryear{Ragot, Martin, and Cojean}{Ragot
  et~al\mbox{.}}{2020}]%
        {ragot}
\bibfield{author}{\bibinfo{person}{Martin Ragot}, \bibinfo{person}{Nicolas
  Martin}, {and} \bibinfo{person}{Salomé Cojean}.}
  \bibinfo{year}{2020}\natexlab{}.
\newblock \showarticletitle{AI-generated vs. Human Artworks. A Perception Bias
  Towards Artificial Intelligence?}
\newblock \bibinfo{journal}{\emph{CHI: Late Breaking Work}}
  (\bibinfo{year}{2020}).
\newblock


\bibitem[\protect\citeauthoryear{Story}{Story}{2020}]%
        {story}
\bibfield{author}{\bibinfo{person}{The~Art Story}.}
  \bibinfo{year}{2020}\natexlab{}.
\newblock \showarticletitle{Ernst Ludwig Kirchner - Biography and Legacy}. In
  \bibinfo{booktitle}{\emph{https://www.theartstory.org/artist/kirchner-ernst-ludwig/life-and-legacy/\#nav}}.
\newblock


\bibitem[\protect\citeauthoryear{Sung}{Sung}{2019}]%
        {sung}
\bibfield{author}{\bibinfo{person}{Morgan Sung}.}
  \bibinfo{year}{2019}\natexlab{}.
\newblock \showarticletitle{The AI Renaissance portrait generator isn't great
  at painting people of color}.
\newblock
  \bibinfo{journal}{\emph{https://mashable.com/article/ai-portrait-generator-pocs/}}
  (\bibinfo{year}{2019}).
\newblock


\bibitem[\protect\citeauthoryear{Tan, Chan, Aguirre, and Tanaka}{Tan
  et~al\mbox{.}}{2017}]%
        {artgan}
\bibfield{author}{\bibinfo{person}{Wei~Ren Tan}, \bibinfo{person}{Chee~Seng
  Chan}, \bibinfo{person}{Hernan Aguirre}, {and} \bibinfo{person}{Kiyoshi
  Tanaka}.} \bibinfo{year}{2017}\natexlab{}.
\newblock \showarticletitle{ArtGAN: Artwork Synthesis with Conditional
  Categorical GANs}.
\newblock \bibinfo{journal}{\emph{ArXiV}} (\bibinfo{year}{2017}).
\newblock


\bibitem[\protect\citeauthoryear{Value}{Value}{2019}]%
        {value}
\bibfield{author}{\bibinfo{person}{Value}.} \bibinfo{year}{2019}\natexlab{}.
\newblock \showarticletitle{AI Artwork Goes Up for Auction for $£30,000$ at
  Sotheby’s}.
\newblock
  \bibinfo{journal}{\emph{https://en.thevalue.com/articles/sothebys-ai-memories-of-passersby}}
  (\bibinfo{year}{2019}).
\newblock


\bibitem[\protect\citeauthoryear{van Gogh}{van Gogh}{1886}]%
        {harrison}
\bibfield{author}{\bibinfo{person}{Vincent van Gogh}.}
  \bibinfo{year}{1886}\natexlab{}.
\newblock \showarticletitle{Letter to Horace M. Livens, Translated by Robert
  Harrison}.
\newblock
  \bibinfo{journal}{\emph{http://www.webexhibits.org/vangogh/letter/17/459a.htm}}
  (\bibinfo{year}{1886}).
\newblock


\bibitem[\protect\citeauthoryear{van Gogh}{van Gogh}{1888}]%
        {johanna}
\bibfield{author}{\bibinfo{person}{Vincent van Gogh}.}
  \bibinfo{year}{1888}\natexlab{}.
\newblock \showarticletitle{Letter to Wilhelmina van Gogh, Translated by
  Translated by Mrs. Johanna van Gogh-Bonger}.
\newblock
  \bibinfo{journal}{\emph{http://www.webexhibits.org/vangogh/letter/18/W04.htm}}
  (\bibinfo{year}{1888}).
\newblock


\bibitem[\protect\citeauthoryear{VanGoghGallery}{VanGoghGallery}{2020}]%
        {gogh}
\bibfield{author}{\bibinfo{person}{VanGoghGallery}.}
  \bibinfo{year}{2020}\natexlab{}.
\newblock \showarticletitle{VINCENT VAN GOGH: POPPIES}.
\newblock
  \bibinfo{journal}{\emph{https://www.vangoghgallery.com/misc/poppies.html}}
  (\bibinfo{year}{2020}).
\newblock


\bibitem[\protect\citeauthoryear{Voorhies)}{Voorhies)}{2004}]%
        {james}
\bibfield{author}{\bibinfo{person}{James Voorhies)}.}
  \bibinfo{year}{2004}\natexlab{}.
\newblock \showarticletitle{Paul Cézanne (1839–1906)}.
\newblock \bibinfo{journal}{\emph{Heilbrunn Timeline of Art History}}
  (\bibinfo{year}{2004}).
\newblock


\bibitem[\protect\citeauthoryear{Wikiart}{Wikiart}{2020}]%
        {wikiart}
\bibfield{author}{\bibinfo{person}{Wikiart}.} \bibinfo{year}{2020}\natexlab{}.
\newblock \showarticletitle{Visual Art Encyclopedia}.
\newblock \bibinfo{journal}{\emph{https://www.wikiart.org}}
  (\bibinfo{year}{2020}).
\newblock


\bibitem[\protect\citeauthoryear{Wikipedia}{Wikipedia}{2020a}]%
        {automation}
\bibfield{author}{\bibinfo{person}{Wikipedia}.}
  \bibinfo{year}{2020}\natexlab{a}.
\newblock \showarticletitle{Automation Bias}.
\newblock
  \bibinfo{journal}{\emph{https://en.wikipedia.org/wiki/Automation\_bias}}
  (\bibinfo{year}{2020}).
\newblock


\bibitem[\protect\citeauthoryear{Wikipedia}{Wikipedia}{2020b}]%
        {imodern}
\bibfield{author}{\bibinfo{person}{Wikipedia}.}
  \bibinfo{year}{2020}\natexlab{b}.
\newblock \showarticletitle{Futurism}.
\newblock \bibinfo{journal}{\emph{https://en.wikipedia.org/wiki/Futurism}}
  (\bibinfo{year}{2020}).
\newblock


\bibitem[\protect\citeauthoryear{Wikipedia}{Wikipedia}{2020c}]%
        {wiki}
\bibfield{author}{\bibinfo{person}{Wikipedia}.}
  \bibinfo{year}{2020}\natexlab{c}.
\newblock \showarticletitle{Generative Art}.
\newblock
  \bibinfo{journal}{\emph{https://en.wikipedia.org/wiki/Generative\_art}}
  (\bibinfo{year}{2020}).
\newblock


\bibitem[\protect\citeauthoryear{Wikipedia}{Wikipedia}{2020d}]%
        {ukiyo}
\bibfield{author}{\bibinfo{person}{Wikipedia}.}
  \bibinfo{year}{2020}\natexlab{d}.
\newblock \showarticletitle{Ukiyo-e}.
\newblock \bibinfo{journal}{\emph{https://en.wikipedia.org/wiki/Ukiyo-e}}
  (\bibinfo{year}{2020}).
\newblock


\bibitem[\protect\citeauthoryear{Wilber, Fang, Jin, Hertzmann, Collomosse, and
  Belongie}{Wilber et~al\mbox{.}}{2017}]%
        {behance}
\bibfield{author}{\bibinfo{person}{Michael~J. Wilber}, \bibinfo{person}{Chen
  Fang}, \bibinfo{person}{Hailin Jin}, \bibinfo{person}{Aaron Hertzmann},
  \bibinfo{person}{John Collomosse}, {and} \bibinfo{person}{Serge Belongie}.}
  \bibinfo{year}{2017}\natexlab{}.
\newblock \showarticletitle{BAM! The Behance Artistic Media Dataset for
  Recognition Beyond Photography}.
\newblock \bibinfo{journal}{\emph{ICCV}} (\bibinfo{year}{2017}).
\newblock


\bibitem[\protect\citeauthoryear{Wooldridge}{Wooldridge}{2009}]%
        {wooldridge}
\bibfield{author}{\bibinfo{person}{Jeffrey~M. Wooldridge}.}
  \bibinfo{year}{2009}\natexlab{}.
\newblock \showarticletitle{Omitted Variable Bias: The Simple Case}. In
  \bibinfo{booktitle}{\emph{Introductory Econometrics: A Modern Approach}}.
\newblock


\bibitem[\protect\citeauthoryear{Young}{Young}{2019}]%
        {tabularasa}
\bibfield{author}{\bibinfo{person}{David Young}.}
  \bibinfo{year}{2019}\natexlab{}.
\newblock \showarticletitle{Tabula Rasa: Rethinking the Intelligence of Machine
  Minds}.
\newblock
  \bibinfo{journal}{\emph{https://medium.com/@dkyy/tabula-rasa-b5f846e60859}}
  (\bibinfo{year}{2019}).
\newblock


\bibitem[\protect\citeauthoryear{Young}{Young}{2001}]%
        {young}
\bibfield{author}{\bibinfo{person}{James~O. Young}.}
  \bibinfo{year}{2001}\natexlab{}.
\newblock \showarticletitle{Art and Knowledge}.
\newblock \bibinfo{journal}{\emph{Routledge, London, UK}}
  (\bibinfo{year}{2001}).
\newblock


\bibitem[\protect\citeauthoryear{Z., T., and et. al}{Z. et~al\mbox{.}}{2009}]%
        {liberman}
\bibfield{author}{\bibinfo{person}{Lieberman Z.}, \bibinfo{person}{Watson T.},
  {and} \bibinfo{person}{Castro~A et. al}.} \bibinfo{year}{2009}\natexlab{}.
\newblock \showarticletitle{Openframeworks}.
\newblock \bibinfo{journal}{\emph{https://goo.gl/41tycE}}
  (\bibinfo{year}{2009}).
\newblock


\bibitem[\protect\citeauthoryear{Zhu, Park, Isola, and Efros}{Zhu
  et~al\mbox{.}}{2017}]%
        {cyclegan}
\bibfield{author}{\bibinfo{person}{Jun-Yan Zhu}, \bibinfo{person}{Taesung
  Park}, \bibinfo{person}{Phillip Isola}, {and} \bibinfo{person}{Alexei~A.
  Efros}.} \bibinfo{year}{2017}\natexlab{}.
\newblock \showarticletitle{Unpaired Image-to-Image Translation using
  Cycle-Consistent Adversarial Networks}.
\newblock \bibinfo{journal}{\emph{ICCV}} (\bibinfo{year}{2017}).
\newblock


\end{thebibliography}

\appendix

\end{document}